\documentclass{article}

\usepackage[margin=1.0in]{geometry}
\usepackage[utf8]{inputenc} 
\usepackage[T1]{fontenc}    
\usepackage{url}            
\usepackage{booktabs}       
\usepackage{amsfonts}       
\usepackage{nicefrac}       
\usepackage{microtype}      
\usepackage{xcolor}
\usepackage{lineno}

\usepackage{xr-hyper}
\makeatletter
\newcommand*{\addFileDependency}[1]{
  \typeout{(#1)}
  \@addtofilelist{#1}
  \IfFileExists{#1}{}{\typeout{No file #1.}}
}
\makeatother
\newcommand*{\myexternaldocument}[1]{%
    \externaldocument{#1}%
    \addFileDependency{#1.tex}%
    \addFileDependency{#1.aux}%
}

\usepackage{graphicx}
\usepackage{biblatex}
\usepackage{algorithm}
\usepackage[noend]{algpseudocode}
\usepackage{amsmath}
\usepackage{amsthm}
\usepackage{enumitem}
\usepackage{authblk}
\algnewcommand{\IIf}[1]{\State\algorithmicif\ #1\ \algorithmicthen}
\algnewcommand{\EndIIf}{\unskip\ \algorithmicend\ \algorithmicif}

\usepackage[utf8]{inputenc} 
\usepackage[T1]{fontenc}    
\usepackage{hyperref}       
\usepackage{url}            
\usepackage{booktabs}       
\usepackage{amsfonts}       
\usepackage{nicefrac}       
\usepackage{microtype}      

\usepackage{lipsum}
\usepackage{amsmath}
\usepackage{xcolor}
\usepackage{graphicx}
\usepackage{cancel}
\usepackage{ulem}
\usepackage{mathtools}

\usepackage{caption}

\usepackage{todonotes}
\usepackage{datenumber}
\usepackage{eqname}
\usepackage{float}
\usepackage{setspace}

\newcommand{\defeq}{\vcentcolon=}

\usepackage[symbol]{footmisc}

\newcommand{\argmax}{\text{argmax}}
\newcommand{\argmin}{\text{argmin}}

\myexternaldocument{supplement}

\title{Improving performance of deep learning models with axiomatic attribution priors and expected gradients}

\author[1,2,*]{Gabriel Erion}
\author[1,2,*]{Joseph D. Janizek}
\author[1,*]{Pascal Sturmfels}
\author[1,3]{Scott M. Lundberg}
\author[1,**]{Su-In Lee}

\affil[1]{{\small Paul G. Allen School of Computer Science and Engineering, University of Washington}}
\affil[2]{{\small Medical Scientist Training Program, University of Washington}}
\affil[3]{{\small Microsoft Research}}

\affil[*]{{\small These authors contributed equally and are ordered alphabetically}}
\affil[**]{{\small Corresponding: suinlee@cs.washington.edu}}

\begin{document}

\setcounter{page}{1}

\captionsetup{font={small}}

\date{}

{\setstretch{1}
\maketitle
}
\noindent \small {\bf One sentence summary:} We introduce a method for using axiomatic feature attributions to encourage a wide variety of desirable behaviors in deep models.
\begin{abstract}
\noindent
Recent research has demonstrated that feature attribution methods for deep networks can themselves be incorporated into training; these \textit{attribution priors} optimize for a model whose attributions have certain desirable properties -- most frequently, that particular features are important or unimportant. These attribution priors are often based on attribution methods that are not guaranteed to satisfy desirable interpretability axioms, such as completeness and implementation invariance.
Here, we introduce
attribution priors to optimize for higher-level properties of explanations, such as smoothness and sparsity, enabled by a fast new attribution method formulation called \textit{expected gradients} that satisfies many important interpretability axioms. This improves model performance on many real-world tasks where previous attribution priors fail.
Our experiments show that the gains from combining higher-level attribution priors with expected gradients attributions are consistent across image, gene expression, and health care data sets. We believe this work motivates and provides the necessary tools to support the widespread adoption of axiomatic attribution priors in many areas of applied machine learning. The implementations and our results have been made freely available to academic communities.\footnote{Packages implementing our attribution priors and expected gradients, as well as experiments illustrating key results, are available at \url{https://github.com/suinleelab/attributionpriors}.}
\end{abstract}

\section{Introduction}

Recent work on interpreting machine learning (ML) models focuses on \textit{feature attribution methods}. Given an input datum, a model, and a prediction, such methods assign a number to each input feature that represents how important the feature was for making the prediction. Current research is also investigating the axioms that attribution methods should satisfy \cite{lundberg2017unified, sundararajan2017axiomatic, vstrumbelj2014explaining, datta2016algorithmic} and how they provide insight into model behavior \cite{lundberglocal, lundberg2018explainable, sayres2019using, zech2018variable}. Feature attribution methods often reveal problems in a model or dataset. For example, a model may place too much importance on undesirable features, rely on many features when sparsity is desired, or be sensitive to high frequency noise. In such cases, humans often have a prior belief about how a model should treat input features but find it difficult to mathematically encode this prior for neural networks in terms of the model parameters.

Ross et al. \cite{ross2017right} introduced the first instance of what we call an \textit{attribution prior}: if it is possible for explanations to reveal problems in a model, constraining the model's explanations during training can help the model avoid such problems. Given a binary indicator of whether each feature should or should not be important for making predictions on each sample in the dataset, their method penalizes the gradients of unimportant features. 
However, two drawbacks limit this method's applicability to real-world problems. First, gradients do not satisfy the same theoretical guarantees as modern feature attribution methods. This leads to well-known problems such as saturation: operations, like ReLUs and sigmoids, which have large flat ``saturated'' regions, can lead to 0 gradient attribution even for important features \cite{sundararajan2017axiomatic}. Second, it can be difficult to specify which features should be important in a binary manner.
A very recent publication successfully uses Ross et al's gradient-based prior as part of a human-in-the-loop strategy to improve model generalization performance and user trust, as well as contributing their own model-agnostic method for penalizing feature importances \cite{schramowski2020making}. Such results create a clear synergy with our study, which improves the quality of calculated importances and develops new forms of attribution priors. This has the potential to greatly expand both the number of ways that a human-in-the-loop can influence deep models and the precision with which they can do so.

More recent work discusses the need for priors that incorporate human intuition in order to develop robust and interpretable models \cite{ilyas2019notbugs}. 
Still, it remains challenging to encode priors such as ``have smoother attributions across an image'' or ``treat this group of features similarly'' by penalizing a model's input gradients or parameters. Some recent attribution priors have proposed regularizing integrated gradients attributions \cite{liu2019incorporating,chen2019robust}. While promising, this work suffers from three major weaknesses: it does not clearly demonstrate improvements over gradient-based attribution priors, it penalizes attribution deviation from a target value rather than encoding sophisticated priors such as those we mention above, and it imposes a large computational cost by training with tens to hundreds of reference samples per batch.

A contemporary method called contextual decomposition explanation penalization (CDEP) uses a framework similar to attribution priors and penalizes explanations generated by the contextual decomposition (CD) method \cite{rieger2020interpretations}. 
CDEP claims to greatly increase computational efficiency compared to the method presented in this paper; however, we demonstrate here that these observed performance differences were actually the result of errors in the re-implementation of the method (Supplement Section \ref{sec:supp_cdep}).
In many cases, our method actually trains faster than CDEP. Our approach also maintains several advantages, including a much more flexible attribution method that works with a wider array of models (any model with gradients) and penalties (any differentiable function of the attributions for each feature). For further discussion of related work, including a discussion of cases where CDEP should work well, see Supplement Section \ref{sec: supp_related_work}.

The main contribution of this work is a broadened interpretation of attribution priors that includes any case in which the training objective incorporates differentiable functions of a model's axiomatic feature attributions. This can be seen as a generalization of gradient-based regularization \cite{lecun2010mnist, ross2017right, yu2018towards, jakubovitz2018improving, roth2018adversarially} and it can be used to encode meaningful domain knowledge more effectively than existing methods. Whereas previous attribution priors generally took the form of ``encourage feature $i$'s attribution to be near a pre-determined target value,'' the priors we present here consider relative importance among \textit{multiple} features and do not require pre-determined target values for any feature's attribution. Specifically, we introduce an \textit{image prior} enforcing that neighboring pixels have similar attributions, a \textit{graph prior} for biological data enforcing that related genes have similar attributions, and a \textit{sparsity prior} enforcing that a few features have large attributions while all others have near-zero ones.

We also introduce a new general-purpose feature attribution method to enforce these priors, \textit{expected gradients}, which eliminates a hyperparameter choice required by integrated gradients \cite{sundararajan2017axiomatic}. Further, it can be efficiently regularized under an attribution prior due to its formulation as an expectation, which naturally lends itself to batched estimates of the attribution. Since these attributions are used not only to interpret trained models, but also as part of the training objective itself, it is essential to guarantee that the attributions will be of high quality. We therefore show that our attribution method satisfies important interpretability axioms.

Across three different prediction tasks, we show that training with expected gradients outperforms training with previous, more limited versions of attribution priors. On images, our image prior produces a model that is more interpretable and generalizes better to noisy data. On gene expression data, our graph prior reduces prediction error and better captures biological signal. Finally, on a patient mortality prediction task, our sparsity prior yields a sparser model and improves performance when learning from limited training data.

\section{Results}
\label{sec: Results}
\subsection{Attribution priors are a flexible framework for encoding domain knowledge.}
Let $X \in \mathbb{R}^{n \times p}$ denote a dataset with labels $y \in \mathbb{R}^{n \times o}$, where $n$ is the number of samples, $p$ is the number of features, and $o$ is the number of outputs. In standard deep learning, we find optimal parameters $\theta$ by minimizing loss, with a regularization term $\Omega'(\theta)$ weighted by $\lambda'$ on the parameters:
\[\theta  = \text{argmin}_\theta \mathcal{L}(\theta;X,y) + \lambda' \Omega' (\theta). \]
Attribution priors involve a model's attributions, represented by the matrix $\Phi(\theta, X)$, where each entry $\phi_i^\ell$ is the importance of feature $i$ in the model's output for sample $\ell$. The attribution prior is a scalar-valued penalty function of the feature attributions $\Omega(\Phi(\theta,X))$, which represents a log-transformed prior probability distribution over possible attributions ($\lambda$ is the regularization strength). The attribution prior is modular and agnostic to the particular attribution method. This results in the optimization:
\[\theta  = \text{argmin}_\theta \mathcal{L}(\theta;X,y) + \lambda \Omega(\Phi(\theta,X)), \]
where the standard regularization term has simply been replaced with an arbitrary, differentiable penalty function on the feature attributions. 

While feature attributions have previously been used in training (more details in \ref{sec:previous_attribution_priors}) \cite{ross2017right,liu2019incorporating}, our approach offers two novel components: First, we demonstrate that calculating $\Phi$ with attribution methods that satisfy previously-established \textit{interpretability axioms} improves performance (see Section 2.2 and \ref{sec:methods_eg} for further discussion of interpretability axioms). Second, rather than simply encouraging each feature's attribution to be near a target value as in previous work, we enforce \textit{high-level} priors over the relationships between features. In image data, we use a Laplace 0-mean prior on the difference between attributions of adjacent pixels, which encourages a low total variation (high smoothness) of attributions:
\[ \Omega_{\textrm{pixel}} (\Phi(\theta,X)) = \sum_\ell \sum_{i,j} |\phi^\ell_{i+1,j}-\phi^\ell_{i,j}| + |\phi^\ell_{i,j+1}-\phi^\ell_{i,j}|. \]
In gene expression data, we use a Gaussian 0-mean prior on the difference between mean absolute attributions $\bar\phi_i$ of functionally 
related genes, which encourages such similar genes to have similar attributions:
\begin{equation*}
    \Omega_{\textrm{graph}}(\Phi(\theta, X)) = \sum_{i,j} W_{i,j} (\bar{\phi}_{i} - \bar{\phi}_{j})^2 = \bar{\phi}^T L_G \bar{\phi},
\end{equation*}
where $W_{i,j}$ is the weight of connection between two genes in a biological graph, and $L_G$ is the graph Laplacian.
Finally, in health data where sparsity is desired, we use a prior on the Gini coefficient of the mean absolute attributions $\bar\phi_i$, which encourages a small number of features to have a large percentage of the total attribution while others are near-zero:
\begin{equation*}
    \Omega_{\textrm{sparse}}(\Phi(\theta, X)) = -\frac{\sum_{i=1}^p \sum_{j=1}^p |\bar\phi_i-\bar\phi_j|}{n\sum_{i=1}^p \bar\phi_i}= -2 G(\bar\phi),
\end{equation*}
where $G$ is the Gini coefficient.
None of these priors require specifying target values for features, and all improve performance over simpler baselines. For more details on our priors see \ref{sec:methods_priors}, and for more details on previous attribution priors, see \ref{sec:previous_attribution_priors}.
\subsection{Expected gradients outperforms other attribution methods.}

\label{sec:eg}
Attribution priors involve using feature attributions not just as a post-hoc analysis method, but as a key part of the training objective. Thus, it is essential to guarantee as much as possible that the attribution method used will produce high-quality attributions. We propose an axiomatic feature attribution method called \textit{expected gradients}, which avoids problems with existing methods and is naturally suited to being incorporated into training. Expected gradients extends the integrated gradients method \cite{sundararajan2017axiomatic}, and like integrated gradients, satisfies a variety of desirable interpretability axioms such as completeness (the feature attributions sum to the output for a given sample) and implementation invariance (the attributions are identical for any of the infinite possible implementations of the same function). Because these methods satisfy completeness, they are not subject to problems with input saturation, while methods such as gradients that do not satisfy completeness will suffer from input saturation. Because these methods satisfy implementation invariance, they are straight-forward to practically apply to any differentiable model, regardless of specific network architectures (see \ref{sec:methods_eg} for an extended discussion of the interpretability axioms satisfied by expected gradients). 

Integrated gradients generates feature attributions by explaining the difference between the model's output for a sample and its output for a \textit{reference} sample. Choosing this reference is difficult. For example, in image tasks, the image of all zeros is often chosen as a baseline, but doing so implies that black pixels will not be highlighted as important. This problem can be solved by integrating gradients over multiple references. However, this requires multiple integrals, is expensive in terms of time and memory, and may be prohibitive to calculate for every sample in every epoch of training. Our method collapses these multiple integrals using sampling to yield accurate attributions with multiple references that can be calculated quickly (here, $x$ is the sample, $x'$ is a reference, and $D$ is the data distribution):
\begin{align*}
    \textrm{ExpectedGradients}_i(x) &\defeq \int_{x'} \textrm{IntegratedGradients}_i(x,x') p_D(x') d x', \\
    &= \int_{x'} \Bigg( (x_i - x_i') \times \int_{\alpha = 0}^1 \frac{\delta f(x' + \alpha (x - x'))}{\delta x_i} d \alpha \Bigg) p_D(x') d x', \\
    &= \mathop{\mathbb{E}}_{x' \sim D, \alpha \sim U(0, 1)} \bigg [ (x_i - x_i')\times \frac{\delta f(x' + \alpha \times(x - x'))}{\delta x_i} \bigg ].
\end{align*}

If the attribution function $\Phi$ in our attribution prior $\Omega(\Phi(\theta, X))$ is integrated gradients, regularizing $\Phi$ would require hundreds of extra gradient calls every training step (the original integrated gradients paper \cite{sundararajan2017axiomatic} recommends 20 to 300 gradient calls to compute attributions). This makes training with integrated gradients prohibitively slow -- in fact,  \cite{liu2019incorporating} find that using integrated gradients can take up to 30 times longer than standard training even when only back-propagating gradients through part of the network. Even more gradient calls would be necessary if multiple references were used, as proposed above. However, most deep learning models today are trained using some variant of batch gradient descent, where the gradient of a loss function is approximated over many training steps using mini-batches of data. We can use a batch training procedure to approximate expected gradients as well. During training, we let $k$ be the number of samples we draw to compute expected gradients for each mini-batch. Remarkably, as small as $k = 1$ suffices to regularize the explanations because of the \textit{averaging effect} of the expectation formulation over many training samples. This choice of $k$ uses every sample in the training set as a reference over the course of an epoch, with only one additional gradient call per training step. This lets us efficiently regularize expected gradients using the entire training dataset as the reference -- as opposed to the single reference in integrated gradients -- with far fewer gradient calls per epoch (see \ref{sec:methods_eg} for more detail).

We compare expected gradients to other feature attribution methods for 18 benchmarks on synthetic data introduced in \cite{lundberglocal} (Table \ref{tab: sample-table}), using the benchmarking software available as part of the SHAP package.\footnote{https://github.com/slundberg/shap} These benchmark metrics evaluate whether each attribution method finds the most important features for a given dataset and model. Expected gradients significantly outperforms the next best feature attribution method ($p=7.2 \times 10^{-5}$, one-tailed Binomial test). 
We provide more details and additional benchmarks in Supplement Section 
\ref{sec:supp_benchmark}.

\definecolor{nicered}{HTML}{cc002f}
\begin{table*}
  \caption{Results from benchmark software on synthetic data with correlated features. Larger numbers mean a better feature attribution method for all metrics. Metric abbreviations are: K (Keep), R (Remove); P (Positive), N (Negative), A (Absolute); M (Mean masking), R (Resample masking), I (Impute masking). For example, KPM corresponds to the ``Keep Positive with Mean masking'' metric. Each model is trained on $900$ samples and tested using $100$ samples. See Supplement Section 
\ref{sec:supp_benchmark} for details.}
  \label{tab: sample-table}
  \centering
  \begin{tabular}{llllllllll}
    \toprule
    Method & KPM & KPR & KPI & KNM & KNR & KNI & KAM & KAR & KAI \\
    \midrule
    Expected Grad. 
    & \textcolor{nicered}{\textbf{3.731}} & \textcolor{nicered}{\textbf{3.800}} & \textcolor{nicered}{\textbf{3.973}} &  \textcolor{nicered}{\textbf{3.615}} &  \textcolor{nicered}{\textbf{3.551}} &  \textcolor{nicered}{\textbf{3.873}} &  \textcolor{nicered}{\textbf{0.906}} &  \textcolor{nicered}{\textbf{0.903}} & 0.919\\
    Integrated Grad. & 3.667 & 3.736 & 3.920 & 3.543 & 3.476 & 3.808 & 0.905 & 0.899 &  \textcolor{nicered}{\textbf{0.920}} \\
    Gradients & 0.096 
    & 0.122
    & 0.099
    & 0.076
    & -0.112
    & 0.052
    & 0.838
    & 0.823
    & 0.887\\
    Random & 0.033 
    & 0.106
    & 0.077
    & -0.012
    & -0.093
    & -0.053
    & 0.593
    & 0.583
    & 0.715\\
    \bottomrule
    \toprule
    Method & RPM & RPR & RPI & RNM & RNR & RNI & RAM & RAR & RAI \\
    \midrule
    Expected Grad. & \textcolor{nicered}{\textbf{3.612}}
    & \textcolor{nicered}{\textbf{3.575}}
    & \textcolor{nicered}{\textbf{3.525}}
    & \textcolor{nicered}{\textbf{3.759}}
    & \textcolor{nicered}{\textbf{3.830}}
    & \textcolor{nicered}{\textbf{3.683}}
    & \textcolor{nicered}{\textbf{0.897}}
    & \textcolor{nicered}{\textbf{0.885}}
    & \textcolor{nicered}{\textbf{0.880}}\\
    Integrated Grad. & 3.539
    & 3.503
    & 3.365
    & 3.687
    & 3.754
    & 3.543
    & 0.872
    & 0.859
    & 0.822\\
    Gradients & 0.035
    & -0.098
    & -0.020
    & 0.110
    & 0.105
    & 0.108
    & 0.729
    & 0.712
    & 0.616\\
    Random & -0.053
    & -0.100
    & -0.106
    & 0.034
    & 0.092
    & 0.111
    & 0.400
    & 0.400
    & 0.275\\
    \bottomrule
  \end{tabular}
\end{table*}

\subsection{A pixel attribution prior improves robustness to image noise.}
\label{sec:pixelresult}
Prior work on interpreting image models focused on creating \textit{pixel attribution maps}, which assign a value to each pixel indicating how important that pixel was for a model's prediction \cite{selvaraju2017grad, sundararajan2017axiomatic}. Attribution maps can be noisy and difficult to understand due to their tendency to highlight seemingly unimportant background pixels, indicating the model may be vulnerable to adversarial attacks \cite{ross2018improving}. Although we may prefer a model with smoother attributions, existing methods only post-process attribution maps but do not change model behavior \cite{smilkov2017smoothgrad, selvaraju2017grad, fong2017interpretable}. Such techniques may not be faithful to the original model \cite{ilyas2019notbugs}. In this section, we describe how we applied our framework to train image models with naturally smoother attributions.

\begin{figure}[!ht]
        \centering
    \includegraphics[width=0.3\linewidth]{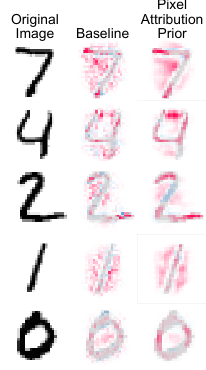}
    \includegraphics[width=0.6\linewidth]{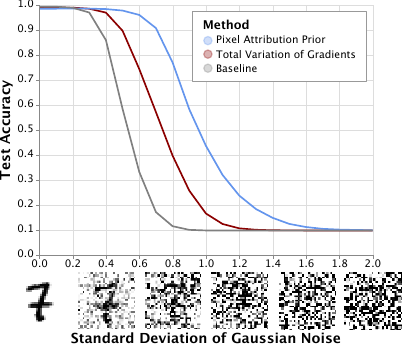}
    \caption{ Left: Expected gradients attributions (from 100 samples) on MNIST for both the baseline and attribution prior models. The latter achieves visually smoother attributions, and it better highlights how the network classifies digits (e.g., the top part of the 4 being very important). Unlike previous methods which take additional steps to smooth saliency maps after training \cite{smilkov2017smoothgrad, fong2017interpretable}, these are \textit{unmodified} saliency maps directly from the learned model. Right: Inducing smoothness in saliency maps leads to robustness to input noise without specifically training for robustness. 
    }
    \label{fig:MNIST}
\end{figure}

\begin{figure*}
        \centering
    \includegraphics[width=0.33\linewidth]{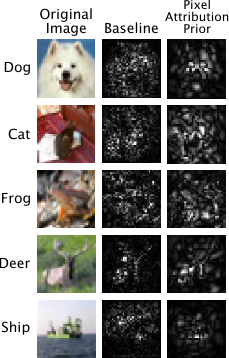}
    \hspace{0.2cm}
    \includegraphics[width=0.57\linewidth]{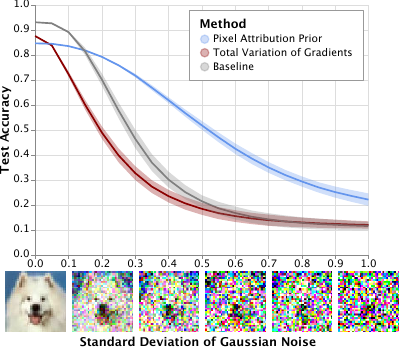}
    \caption{ Left: Expected gradients attributions (from 100 samples) on CIFAR10 for both the baseline model and the model trained with an attribution prior for five randomly selected images classified correctly by both models. Training with an attribution prior generates visually smoother attribution maps in all cases. Notably, these smoothed attributions also appear more localized towards the object of interest. Right: Training with an attribution prior induces robustness to Gaussian noise, achieving more than double the accuracy of the baseline at high noise levels. This robustness is not achievable by choosing gradients as the attribution function. }
    \label{fig:CIFAR10}
\end{figure*}

To regularize pixel-level attributions, we used the following intuition: neighboring pixels should have a similar impact on an image model's output. To encode this intuition, we chose a total variation loss on pixel-level attributions (see \ref{sec:methods_priors} for more detail).
We applied this pixel smoothness attribution prior to the MNIST and CIFAR-10 datasets \cite{lecun2010mnist, krizhevsky2009learning}. On MNIST we trained a two-layer convolutional neural network; for CIFAR-10 we trained a VGG16 network from scratch (see \ref{sec:methodsImageModel} for more details) \cite{simonyan2014very}. In both cases we optimized hyperparameters for the baseline model without an attribution prior. To choose $\lambda$, we searched over values in $[10^{-20}, 10^{-1}]$ and chose the $\lambda$ that minimized the attribution prior penalty and achieved a test accuracy within $10\%$ of the baseline model. Figures \ref{fig:MNIST} and \ref{fig:CIFAR10} display expected gradients attribution maps for both the baseline and the model regularized with an attribution prior on 5 randomly selected test images on MNIST and CIFAR-10, respectively. In all examples, the attribution prior yields a model with visually smoother attributions. Remarkably, in many instances smoother attributions better highlight the target object's structure.

Recent work has suggested that image classifiers are brittle to small domain shifts: small changes in the underlying distribution of the training and test set can significantly reduce test accuracy \cite{recht2019imagenet}. To simulate a domain shift, we applied Gaussian noise to images in the test set and re-evaluated the performance of the regularized and baseline models. As an adaptation of \cite{ross2017right}, we also compared the attribution prior model to regularizing the total variation of gradients with the same criteria for choosing $\lambda$. For each method, we trained 5 models with different random initializations. In Figures \ref{fig:MNIST} and \ref{fig:CIFAR10}, we plot the mean and standard deviation of test accuracy on MNIST and CIFAR-10, respectively, as a function of standard deviation of added Gaussian noise. The figures show that our regularized model is more robust to noise than both the baseline and gradient-based models.

Both the robustness and more intuitive saliency maps our method provides come at the cost of reduced test set accuracy ($0.93 \pm 0.002$ for the baseline vs. $0.85 \pm 0.003$ for pixel attribution prior model on CIFAR-10). The trade-off between robustness and accuracy that we observe is consistent with previous work that suggests image classifiers trained solely to maximize test accuracy rely on features that are brittle and difficult to interpret \cite{ilyas2019notbugs, tsipras2018robustness, zhang2019theoretically}. Despite this trade-off, we find that at a stricter hyperparameter cutoff for $\lambda$ -- within $1\%$ test accuracy of the baseline, rather than $10\%$ -- our methods still achieve modest but significant robustness relative to the baseline. For results at different hyperparameter thresholds, more details on our training procedure, and additional experiments on MNIST, CIFAR-10 and ImageNet, see Supplement Sections \ref{sec:supp_imagenet}, \ref{sec:ImageNetExperiments}, \ref{sec:supp_cifar} and \ref{sec:supp_mnist}.

\subsection{A Graph attribution prior improves anti-cancer drug response prediction.}

In the image domain, our attribution prior took the form of a penalty encouraging smoothness over adjacent pixels. In other domains, there may be prior information about specific relationships between features that can be encoded as a graph (such as social networks, knowledge graphs, or protein-protein interactions). For example, prior work in bioinformatics has shown that protein-protein interaction networks contain valuable information for improving performance on biological prediction tasks \cite{10.1093/bioinformatics/btu293}. Therefore, in this domain we regularized attributions to be smooth over the protein-protein feature graph analogously to the regular graph of pixels in the image.

Incorporating the $\Omega_{\textrm{graph}}$ attribution prior not only led to a model with more reasonable attributions but also improved predictive performance by letting us incorporate prior biological knowledge into the training process. We downloaded publicly available gene expression and drug response data for patients with acute myeloid leukemia (AML, a type of blood cancer) and tried to predict patients' drug response from their gene expression \cite{tyner2018functional}. For this regression task, an input sample was a patient's gene expression profile plus a one-hot encoded vector indicating which drug was tested in that patient, while the label we tried to predict was drug response (measured by IC50, the concentration of the drug required to kill half of the patient's tumor cells). To define the graph used by our prior, we downloaded the tissue-specific gene interaction graph for the tissue most closely related to AML in the HumanBase database \cite{greene2015understanding}.

\begin{figure*}
    \centering
    \includegraphics[width=\linewidth]{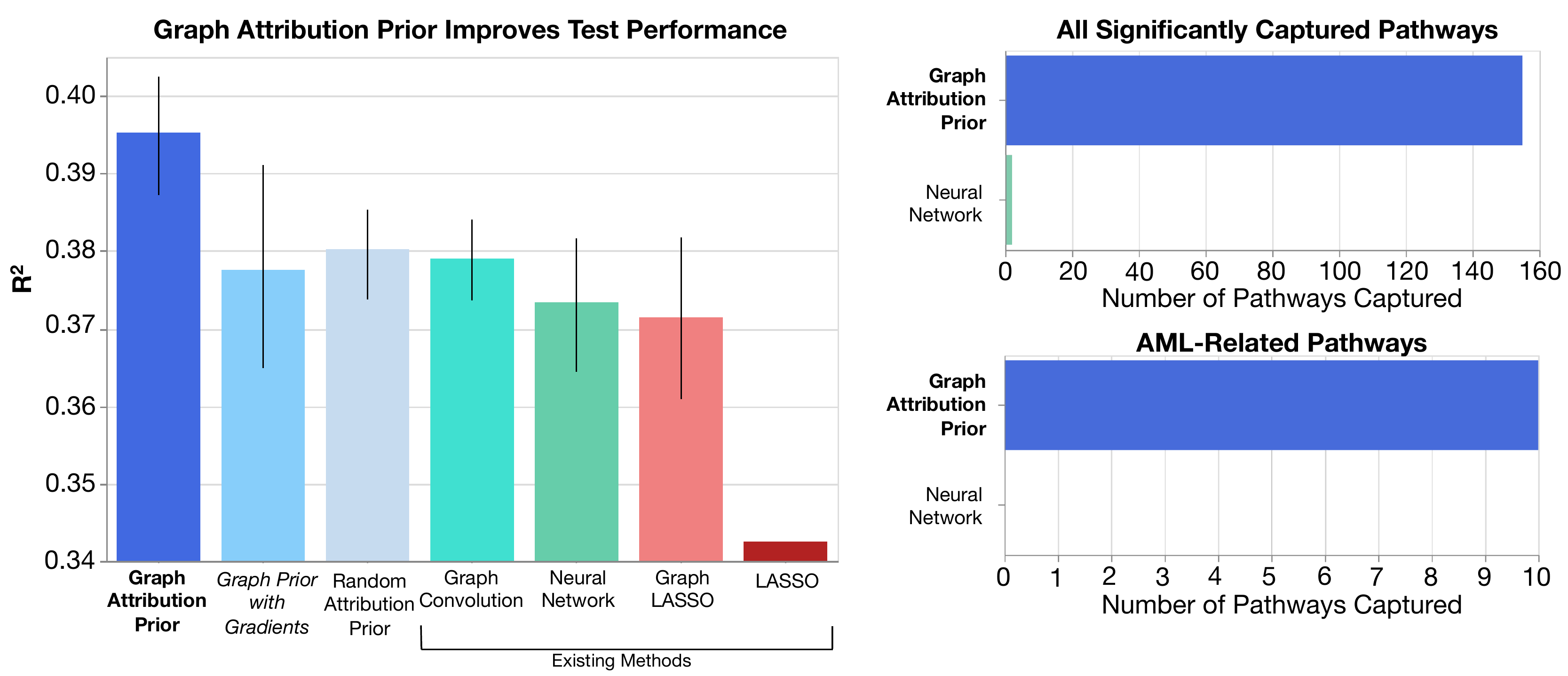}
    \caption{Left: A neural network trained with our graph attribution prior (bold) attains the best test performance, while one trained with the same graph penalty on the gradients (italics, adapted from \cite{ross2017right}) does not perform significantly better than a standard neural network (bars indicate 95\% confidence intervals from bootstrapping the data). Right: A neural network trained with our graph attribution prior captures more biological and AML-relevant pathways than a standard neural network.}
    \label{graphMLP}
\end{figure*}

A two-layer neural network trained with our graph attribution prior ($\Omega_{\textrm{graph}}$) significantly outperforms all other methods in terms of test set performance as measured by $R^2$ (Figure~\ref{graphMLP}, see Methods for significance testing). Unsurprisingly, when we replace the biological graph from HumanBase with a randomized graph, we find that the test performance is no better than the performance of a neural network trained without \textit{any} attribution prior. Extending the method proposed in \cite{ross2017right} by applying our new graph prior as a penalty on the model's \textit{gradients}, rather than a penalty on the axiomatically correct expected gradient feature attribution, does not perform significantly better than a baseline neural network. We also observe substantially improved test performance when using the prior graph information to regularize a linear LASSO model. Finally, we note that our graph attribution prior neural network significantly outperforms graph convolutional neural networks, a recent method for utilizing graph information in deep neural networks \cite{DBLP:journals/corr/KipfW16}.

To find out if our model's attributions match biological intuition, we conducted Gene Set Enrichment Analysis (a modified Kolmogorov–Smirnov test). We measured whether our top genes, ranked by mean absolute feature attribution, were enriched for membership in any pathways (see \ref{sec:biological_experiments} and Supplementary Section \ref{sec:supp_biological} for more detail, including the top pathways for each model) \cite{Subramanian2005}. We find that the neural network with the tissue-specific graph attribution prior captures far more biologically-relevant pathways (increased number of significant pathways after FDR correction) than a neural network without attribution priors (see Figure~\ref{graphMLP}) \cite{Benjamini1995}. Furthermore, the pathways our model uses more closely match biological expert knowledge, i.e., they included prognostically useful AML gene expression profiles as well as important AML-related transcription factors (see Figure~\ref{graphMLP} and Supplementary Section \ref{sec:supp_biological}) \cite{Liu2017,doi:10.1056/NEJMoa040465}. These results are expected, given that neural networks trained without priors can learn a relatively sparse basis of genes that will not enrich for specific pathways (e.g. a single gene from each correlated pathway), while those trained with our graph prior will spread credit among genes. This demonstrates the graph prior's value as an accurate and efficient way to encourage neural networks to treat functionally-related genes similarly.

\subsection{A sparsity prior improves performance with limited training data.}
\textit{Feature selection} and \textit{sparsity} are popular ways to alleviate the curse of dimensionality, facilitate interpretability, 
and improve generalization by building models that use a small number of input features.
A straightforward way to build a sparse deep model is to apply an L1 penalty to the first layer (and possibly subsequent layers) of the network. Similarly, the Sparse Group Lasso (SGL) method penalizes all weights connected to a given feature \cite{feng2017sparse,scardapane2017group}, while a simple existing attribution prior approach \cite{ross2017neural} penalizes the gradients of each feature in the model. 

These approaches suffer from two problems. First, a feature with small gradients or first-layer weights may still strongly affect the model's output \cite{shrikumar2017learning}. A feature whose attribution value (e.g., integrated or expected gradient) is zero
 is much less likely to have any effect on predictions. Second, successfully minimizing the L1 or SGL penalty is not necessarily the best way to create a sparse model. A model that puts weight $w$ on 1 feature is penalized more than one that puts weight $\frac{w}{2p}$ on each of $p$ features. Prior work on sparse linear regression has shown that the Gini coefficient $G$ of the weights, proportional to 0.5 minus the area under the CDF of sorted values, avoids such problems and corresponds more directly to a sparse model \cite{hurley2009comparing,zonoobi2011gini}. We extend this analysis to deep models by noting that the Gini coefficient 
can be written differentiably and used as an attribution prior. 

Here, we show that the $\Omega_{\textrm{sparse}}$ attribution prior can build sparser models that perform better in settings with limited training data. We use a publicly available healthcare mortality prediction dataset of 13,000 patients \cite{miller1973plan}, whose 35 features (118 after one-hot encoding) represent medical data such as a patient's age, vital signs, and laboratory measurements. The binary outcome is survival after 10 years. Sparse models in this setting may enable accurate models to be trained with very few labeled patient samples or reduce cost by accurately risk-stratifying patients using few lab tests. {We randomly sampled training and validation sets of only 100 patients each, placing all other patients in the test set, and ran each experiment 200 times with a new random sample to average out variance.} We built 3-layer binary classifier neural networks regularized using L1, SGL, and sparse attribution prior penalties to predict patient survival{, as well as an L1 penalty on gradients adapted for global sparsity from \cite{ross2017right,ross2017neural}}.  The regularization strength was tuned from $10^{-7}$ to $10^{5}$ using the validation set for all methods (see \ref{sec:sparsityExperiments} and Supplement Section \ref{sec:sparseparams}).

\begin{figure*}
    \centering
    \includegraphics[width=\linewidth]{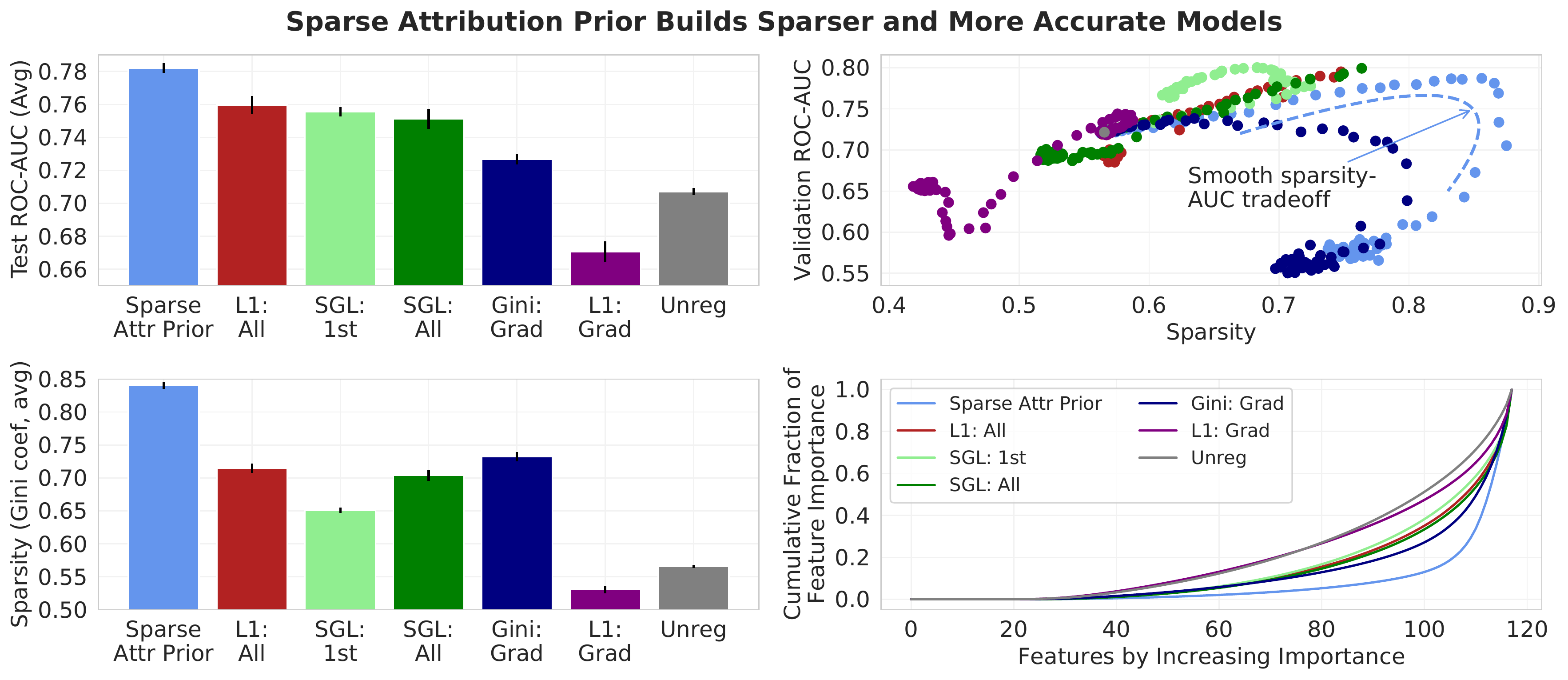}
    \caption{
    Left: A sparse attribution prior enables more accurate test predictions (top) and sparser models (bottom) across 200 small subsampled datasets ($100$ training and $100$ validation samples, all other samples used for test set) than other penalties, including gradients. Top right: Across the full range of tuned parameters, the sparse attribution prior achieves the greatest sparsity and a smooth sparsity-validation performance trade-off. Bottom right: A sparse attribution prior concentrates a larger fraction of global feature importance in the top few features. 
    ``Gini'', ``L1'', and ``SGL'' indicate the Gini, L1, and SGL penalties respectively. ``Grad'' indicates a penalty on the gradients, ``All'' indicates a penalty on all weights in the model, and ``1st'' indicates a penalty on only the first weight layer.}
    \label{fig:nhanes_reg}
\end{figure*}

The sparse attribution prior enables more accurate test predictions and sparser models when limited training data is available (Figure \ref{fig:nhanes_reg}), with $p<10^{-4}$ and $T\geq4.314$ by paired-samples $T$-test for all comparisons. 
We also plot the average cumulative importance of sorted features and find that the sparse attribution prior more effectively concentrates importance in the top few features. In particular, we observe that L1 penalizing the model's gradients as in \cite{ross2017neural} rather than its expected gradients attributions performs poorly in terms of both sparsity and performance. A Gini penalty on gradients improves sparsity but does not outperform other baselines like SGL and L1 in ROC-AUC.
Finally, we plot the average sparsity of the models (Gini coefficient) against their validation ROC-AUC across the full range of regularization strengths. The sparse attribution prior exhibits higher sparsity than other models and a smooth tradeoff between sparsity and ROC-AUC. 
Details and results for other penalties, including L2, dropout, and other attribution priors, are in Supplement Section \ref{sec:supp_sparsity}.

\section{Discussion}

The immense popularity of deep learning has driven its application in many areas with diverse, complicated domain knowledge. While it is in principle possible to hand-design network architectures to encode this knowledge, a more practical approach involves the use of attribution priors, which penalize the importance a model places on each of its input features when making predictions. Unfortunately, previous attribution priors have been limited, both theoretically and computationally. Binary penalties only specify whether features should or should not be important and fail to capture relationships among features. Approaches that only focus on a model's input gradients change the local decision boundary but often fail to impact a model's underlying decision-making. Attribution priors on more complicated attributions, like integrated gradients, have proven computationally difficult. 

Our work advances previous work both by introducing novel, flexible attribution priors for multiple domains and by enabling the training of such priors with a newly defined feature attribution method. Our priors lead to smoother and more interpretable image models, biological predictive models that incorporate graph-based prior knowledge, and sparser healthcare models that perform better in data-scarce scenarios. Our attribution method not only enables the training of said priors, but also outperforms its predecessor -- integrated gradients -- in terms of reliably identifying the features models use to make predictions. 

There remain many avenues for future work in this area. We chose to base our prior on an improved version of integrated gradients because it is the most prominent differentiable feature attribution method we are aware of, but a wide array of other attribution methods exist. Our framework makes it straightforward to substitute any other attribution method as long as it is differentiable, and studying the effectiveness of other attribution methods as priors would be valuable. In addition, while we develop new, more sophisticated attribution priors and show their value, there is ample room to improve on our priors and evaluate entirely new ones for other tasks. Determining the best attribution priors for particular tasks opens a further avenue of research. We believe that surveys of domain experts to establish model desiderata for particular applications will help to develop the best priors for any given situation while offering a valuable opportunity to put humans in the loop. 
Overall, the dual advances of sophisticated attribution priors and expected gradients enable a broader view of attribution priors: as tools to achieve domain-specific goals without sacrificing efficiency. 

\newpage
\clearpage
\section*{\hfil Methods \hfil}
\setcounter{section}{0}
\renewcommand*{\thesection}{Methods~\arabic{section}}
\renewcommand*{\thesubsection}{Methods~\arabic{section}.\arabic{subsection}}

\section{Previous attribution priors}
\label{sec:previous_attribution_priors}
The first instance of what we now call an attribution prior was proposed in \cite{ross2017right}, where the regularization term was modified to place a constant penalty on the gradients of undesirable features:
\[ \theta  = \text{argmin}_\theta \mathcal{L}(\theta;X,y) + \lambda''||A \odot  \frac{\partial\mathcal{L}}{\partial X}||^2_F . \]
Here, the attribution method is the gradients of the model, represented by the matrix $\frac{\partial\mathcal{L}}{\partial X}$ whose $\ell, i$th entry is the gradient of the loss at the $\ell$th sample with respect to the $i$th feature. $A$ is a binary matrix indicating which features should be penalized in which samples.

A more general interpretation of attribution priors is that \textit{any function} of \textit{any feature attribution method} could be used to penalize a loss function, thus encoding prior knowledge about what properties the attributions of a model should have. For some model parameters $\theta$, let $\Phi(\theta, X)$ be a feature attribution method, which is a function of $\theta$ and the data $X$. Let $\phi_i^\ell$ be the feature importance of feature $i$ in sample $\ell$. We formally define an \textit{attribution prior} as a scalar-valued penalty function of the feature attributions $\Omega(\Phi(\theta,X))$, which represents a log-transformed prior probability distribution over possible attributions: 
\[\theta  = \text{argmin}_\theta \mathcal{L}(\theta;X,y) + \lambda \Omega(\Phi(\theta,X)), \] where $\lambda$ is the regularization strength. Note that the attribution prior function $\Omega$ is agnostic to the attribution method $\Phi$. 

Previous attribution priors \cite{ross2017right, liu2019incorporating} required specifying an exact target value for the model's attributions, but often we do not know in advance which features are important in advance. In general, there is no requirement that $\Phi(\theta,X)$ constrain attributions to particular values. Section \ref{sec: Results} presented three newly developed attribution priors for different tasks that improve performance without requiring pre-specified attribution targets for any particular feature.

\section{Expected gradients}
\label{sec:methods_eg}

Expected gradients is an extension of integrated gradients \cite{sundararajan2017axiomatic} with fewer hyperparameter choices. Like several other attribution methods, integrated gradients aims to explain the difference between a model's current prediction and the prediction that the model would make when given a baseline input. This baseline input is meant to represent some uninformative reference input that represents not knowing the value of the input features. Although choosing such an input is necessary for several feature attribution methods \cite{sundararajan2017axiomatic, shrikumar2017learning, binder2016layer}, the choice is often made arbitrarily. For example, for image tasks, the image of all zeros is often chosen as a baseline, but doing so implies that black pixels will not be highlighted as important by existing feature attribution methods. In many domains, it is not clear how to choose a baseline that correctly represents a lack of information.

Our method avoids an arbitrary choice of baseline; it models not knowing the value of a feature by integrating over a dataset. For a model $f$, the \textit{integrated gradients} value for feature $i$ is defined as: 
\[
    \textrm{IntegratedGradients}_i(x, x') \defeq (x_i - x_i') \times \int_{\alpha = 0}^1 \frac{\delta f(x' + \alpha (x - x'))}{\delta x_i} d \alpha, 
\]
where $x$ is the target input and $x'$ is baseline input.
To avoid specifying $x'$, we define the \textit{expected gradients} value for feature $i$ as:
\[
    \textrm{ExpectedGradients}_i(x) \defeq \int_{x'} \Bigg( (x_i - x_i') \times \int_{\alpha = 0}^1 \frac{\delta f(x' + \alpha (x - x'))}{\delta x_i} d \alpha \Bigg) p_D(x') d x',
\] where $D$ is the underlying data distribution. Since expected gradients is also a diagonal path method, it satisfies the same axioms as integrated gradients \cite{friedman2004paths}. Directly integrating over the training distribution is intractable; therefore, we instead reformulate the integrals as expectations:
\[
    \textrm{ExpectedGradients}_i(x) \defeq  \\ \mathop{\mathbb{E}}_{x' \sim D, \alpha \sim U(0, 1)} \bigg [ (x_i - x_i')\times \frac{\delta f(x' + \alpha \times(x - x'))}{\delta x_i} \bigg ].
\]

This expectation-based formulation lends itself to a natural, sampling based approximation method: (1) draw samples of $x'$ from the training dataset and $\alpha$ from $U(0, 1)$, (2) compute the value inside the expectation for each sample and (3) average over samples.

Expected gradients also satisfies a set of important interpretability axioms: implementation invariance, sensitivity, completeness, linearity, and symmetry-preserving.
\begin{itemize}
\item \textit{Implementation invariance} states that two networks with outputs that are equal over all inputs should have equivalent attributions. Any attribution method based on the gradients of a network will satisfy this axiom \cite{sundararajan2017axiomatic}, meaning that integrated gradients, expected gradients, and gradients will all be implementation invariant. 
\item \textit{Sensitivity} (sometimes called Dummy) states that when a model does not depend on a feature at all, it receives zero importance. Integrated gradients, expected gradients, and gradients all satisfy sensitivity because the gradient w.r.t. an irrelevant feature will be 0 everywhere. 
\item \textit{Completeness} states that the attributions should sum to the difference between the output of a function at the input to be explained and the output of that function at a baseline. Gradients do \textit{not} satisfy completeness due to saturation at the inputs; elements like ReLUs may cause gradients to be zero, making completeness impossible \cite{sundararajan2017axiomatic}. Integrated gradients and expected gradients both satisfy completeness due to the gradient theorem (fundamental theorem of calculus for line integrals) \cite{sundararajan2017axiomatic}. For expected gradients, the function being integrated is the expectation of the model's output, so completeness means that the attributions sum to the difference between the model's output for the input and the model's output averaged over all possible baselines. 
\item \textit{Linearity} states that for a model that is a linear combination of two submodels $f(x)=af_1(x)+bf_2(x)$, the attributions are a linear combination of the submodels' attributions $\phi(x)=a\phi_1(x)+b\phi_2(x)$. This will hold for integrated gradients, expected gradients, and gradients because gradients are linear.
\item \textit{Symmetry-preserving} states that symmetric variables with identical values should achieve identical attributions. Integrated gradients is symmetry preserving since it is a straight line path method, and expected gradients will also be symmetry preserving, as a symmetric function of symmetric functions will itself be symmetrical \cite{sundararajan2017axiomatic}.
\end{itemize}

The formulation of expected gradients enables an order of magnitude increase in computational efficiency relative to previous approaches for training with attribution priors. We further improve performance by reducing the need for additional data reading. Specifically, for each input in a batch of inputs, we need $k$ additional inputs to calculate expected gradients for that input batch. As long as $k$ is smaller than the batch size, we can avoid any additional data reading by re-using the same batch of input data as a reference batch, as in \cite{zhang2017mixup}. We accomplish this by shifting the batch of input $k$ times, such that each input in the batch uses $k$ other inputs from the batch as its reference values.

\section{Specific priors}
\label{sec:methods_priors}

Here, we elaborate on the explicit form of the attribution priors we used in this paper. In general, minimizing the error of a model corresponds to maximizing the likelihood of the data under a generative model consisting of the learned model plus parametric noise. For example, minimizing mean squared error in a regression task corresponds to maximizing the likelihood of the data under the learned model, assuming Gaussian-distributed errors:
\[\argmin_\theta ||f_\theta(X)-y||_2^2 = \argmax_\theta \text{exp}(-||f_\theta(X)-y||_2^2)=\theta_{MLE},\]
where $\theta_{MLE}$ is the maximum-likelihood estimate of $\theta$ under the model $Y=f_\theta(X)+\mathcal{N}(0,\sigma)$.

An additive regularization term is equivalent to adding a multiplicative (independent) prior to yield a maximum a posteriori estimate:
\[\argmin_\theta ||f_\theta(X)-y||_2^2 + \lambda||\theta||_2^2= \argmax_\theta \text{exp}(-||f_\theta(X)-y||_2^2)\text{exp}(-\lambda||\theta||_2^2)=\theta_{MAP},\]
Here, adding an L2 penalty is equivalent to MAP for $Y=f_\theta(X)+\mathcal{N}(0,\sigma)$ with a $\mathcal{N}(0,\frac{1}{\lambda})$ prior.
We next discuss the functional form of the attribution priors enforced by our penalties. 

\subsection{Pixel attribution prior}

Our pixel attribution prior is based on the anisotropic total variation loss and is given as follows: 
\[ \Omega_{\textrm{pixel}} (\Phi(\theta,X)) = \sum_\ell \sum_{i,j} |\phi^\ell_{i+1,j}-\phi^\ell_{i,j}| + |\phi^\ell_{i,j+1}-\phi^\ell_{i,j}|, \]
where $\phi^\ell_{i, j}$ is the attribution for the $i, j$-th pixel in the $\ell$-th training image. Research shows \cite{bardsley2012laplace} that this penalty is equivalent to placing 0-mean, iid, Laplace-distributed priors on the differences between adjacent pixel values, i.e., $\phi^\ell_{i+1,j}-\phi^\ell_{i,j} \sim \text{Laplace}(0,\lambda^{-1})$ and $\phi^\ell_{i,j+1}-\phi^\ell_{i,j} \sim \text{Laplace}(0,\lambda^{-1})$.
\cite{bardsley2012laplace} does not call our penalty ``total variation,'' but it is in fact the widely used anisotropic version of total variation and is directly implemented in Tensorflow \cite{abadi2016tensorflow,lou2015weighted,
shi2013efficient}.

\subsection{Graph attribution prior}
For our graph attribution prior, we used a protein-protein or gene-gene interaction network and represented these networks as a weighted, undirected graph. Formally, assume we have a weighted adjacency matrix $W \in \mathbb{R}_+^{p \times p}$ for an undirected graph, where the entries encode our prior belief about the pairwise similarity of the importances between two features. For a biological network, $W_{i,j}$ encodes either the probability or strength of interaction between the $i$-th and $j$-th genes (or proteins). We encouraged similarity along graph edges by penalizing the squared Euclidean distance between each pair of feature attributions in proportion to how similar we believe them to be. Using the graph Laplacian ($L_G = D - W$), where $D$ is the diagonal degree matrix of the weighted graph, this becomes:

\begin{equation*}
    \Omega_{\textrm{graph}}(\Phi(\theta, X)) = \sum_{i,j} W_{i,j} (\bar{\phi}_{i} - \bar{\phi}_{j})^2 = \bar{\phi}^T L_G \bar{\phi}.
\end{equation*}

In this case, we choose to penalize \textit{global} rather than local feature attributions. We define $\bar{\phi}_i$ to be the importance of feature $i$ across all samples in our dataset, where this global attribution is calculated as the average magnitude of the feature attribution across all samples: $\bar{\phi}_i = \frac{1}{n} \sum_{\ell = 1}^{n} |\phi_i^{\ell}| $. Just as the image penalty is equivalent to placing a Laplace prior on adjacent pixels in a regular graph, the graph penalty $\Omega_{\textrm{graph}}$ is equivalent to placing a Gaussian prior on adjacent features in an arbitrary graph with Laplacian $L_G$ \cite{bardsley2012laplace}.

\subsection{Sparse attribution prior}
Our sparsity prior uses the Gini coefficient $G$ as a penalty, which is written:
\begin{equation*}
    \Omega_{\textrm{sparse}}(\Phi(\theta, X)) = -\frac{\sum_{i=1}^p \sum_{j=1}^p |\bar\phi_i-\bar\phi_j|}{n\sum_{i=1}^p \bar\phi_i}= -2 G(\bar\phi),
\end{equation*}
By taking exponentials of this function, we find that minimizing the sparsity regularizer is equivalent to maximizing likelihood under a prior proportional to the following:
\[ \prod_{i=1}^p \prod_{j=1}^p \text{exp} \left(\frac{1}{\sum_{i=1}^p \bar\phi_i} |\bar\phi_i - \bar\phi_j| \right), \]
To our knowledge, this prior does not directly correspond to a named distribution. However, we observe that its maximum value occurs when one $\bar\phi_i$ is 1 and all others are $0$, and that its minimum occurs when all $\bar\phi_i$ are equal. This is similar to the total variation penalty $\Omega_\textrm{image}$, but it is normalized and has a flipped sign to \textit{encourage} differences. The corresponding attribution prior is maximized when global attributions are zero for all but one feature and minimized when attributions are uniform across features.

\section{Image model experimental settings}
\label{sec:methodsImageModel}

We trained a VGG16 model from scratch modified for the CIFAR-10 dataset, containing 60,000 colored 32x32-pixel images divided into 10 categories, as in \cite{liu2015very}. To train this network, we used stochastic gradient descent with an initial learning rate of 0.1 and an exponential decay of 0.5 applied every 20 epochs. Additionally, we used a momentum level of 0.9. For augmentation, we shifted each image horizontally and vertically by a pixel shift uniformly drawn from the range [-3, 3], and we randomly rotated each image by an angle uniformly drawn from the range [-15, 15]. We used a batch size of 128. Before training, we normalized the training dataset to have zero mean and unit variance, and standardized the test set with the mean and variance of the training set. We used $k = 1$ background reference samples for our attribution prior while training. When training with attributions over images, we first normalized the per-pixel attribution maps by dividing by the standard deviation before computing the total variation; otherwise, the total variation can be made arbitrarily small without changing model predictions by scaling down the pixel attributions close to 0. See Supplement Section \ref{sec:supp_cifar} for more details.

We repeated the same experiment as above on MNIST, which contains 60,000 black-and-white 28x28-pixel images of handwritten digits. We trained a CNN with two convolutional layers and a single hidden layer. The convolutional layers each had 5x5 filters, a stride length of 1, and 32 and 64 filters total. Each convolutional layer was followed by a max pooling layer of size 2 with stride length 2. The hidden layer had 1024 units and a dropout rate of 0.5 during training \cite{srivastava2014dropout}. Dropout was turned off when calculating the gradients with respect to the attributions. We trained with the Adam optimizer with the default parameters  ($\alpha=0.001, \beta_1=0.9, \beta_2=0.999, \epsilon=10^{-8}$) \cite{kingma2014adam}. We trained with an initial learning rate of 0.0001, with an exponential decay of 0.95 for every epoch, for a total of 60 epochs. For all models, we trained with a batch size of 50 images and used $k=1$ background reference sample per attribution while training. See Supplement Section \ref{sec:supp_mnist} for more details.

\section{Biological experiments}
\label{sec:biological_experiments}

\subsection{Significance testing of results}

To test the difference in $R^2$ attained by each method, we used a T-test for the means of two independent samples of scores (as implemented in SciPy) \cite{2020SciPy-NMeth}. This is a two-sided test and can be applied to $R^2$ since $R^2$ is a linear transformation of mean squared error, which satisfies normality assumptions by the central limit theorem. When we compare the $R^2$ attained from 10 independent retrainings of the neural network to the $R^2$ attained from 10 independent retrainings of the attribution prior model, we find that predictive performance is significantly higher for the model with the graph attribution prior (t-statistic $= 3.59$, $p = 2.06 \times 10^{-3}$).

To ensure that the increased performance in the attribution prior model was due to real biological information, we replaced the gene-interaction graph with a randomized graph (symmetric matrix with identical number of non-zero entries to the real graph, but entries placed in random positions). We then compared the $R^2$ attained from 10 independent retrainings of a neural network with no graph attribution prior to 10 independent retrainings of an neural network regularized with the random graph and found that test error was not significantly different between these two models (t-statistic $= 1.25$,$p = 0.23$). We also compared to graph convolutional neural networks, and found that our network with a graph attribution prior outperformed the graph convolutional neural network (t-statistic $= 3.30$,$p = 4.0 \times 10^{-3}$).

\subsection{Train/validation/test set allocation}

To increase the number of samples in our dataset, we used as a feature the identity of the drug being tested, rather than one of a number of possible output tasks in a multi-task prediction. This follows from prior literature on training neural networks to predict drug response \cite{doi:10.1093/bioinformatics/btx806}. This yielded 30,816 samples (covering 218 patients and 145 anti-cancer drugs). Defining a sample as a drug and a patient, however, meant we had to choose carefully how to stratify samples into our train, validation, and test sets. While it is perfectly legitimate in general to randomly stratify samples into these sets, we wanted to specifically focus on how well our model could learn trends from gene expression data that would generalize to new patients. Therefore, we stratified samples at a patient-level rather than at the level of individual samples (e.g., no samples from any patient in the test set ever appeared in the training set). We split 20\% of the total patients into a test set (6,155 samples) and then split 20\% of the training data into a validation set for hyperparameter selection (4,709 samples).

\subsection{Model class implementations and hyperparameters tested}

\hspace{10pt} \textbf{LASSO.} We used the scikit-learn implementation of the LASSO \cite{Tibshirani1996,scikit-learn}. We tested a range of $\alpha$ parameters from $10^{-9}$ to $1$, and we found that the optimal value for $\alpha$ was $10^{-2}$ by mean squared error on the validation set. 

\textbf{Graph LASSO.} For our Graph LASSO, we used the Adam optimizer in TensorFlow \cite{abadi2016tensorflow}, with a learning rate of $10^{-5}$ to optimize the following loss function:

\begin{equation}
    \mathcal{L}(w;X,y) = \| Xw - y  \|_2^2 + \lambda'\|w\|_1 + \nu' w^T L_G w,
\end{equation}
where $w \in \mathbb{R}^d$ is the weights vector of our linear model and $L_G$ is the graph Laplacian of our HumanBase network \cite{greene2015understanding}. In particular, we downloaded the ``Top Edges'' version of the hematopoietic stem cell network, which was thresholded to only have non-zero values for pairwise interactions that had a posterior probability greater than $0.1$. We used the value of $\lambda'$ selected as optimal in the regular LASSO model ($10^{-2}$, which corresponds to the $\alpha$ parameter in scikit-learn) and then tuned over $\nu'$ values ranging from $10^{-3}$ to $100$. We found that a value of $10$ was optimal according to MSE on the validation set.

\textbf{Neural networks.} We tested a variety of hyperparameter settings and network architectures via validation set performance to choose our best neural networks, including the following feed-forward network architectures (where each element in a list denotes the size of a hidden layer): [512,256], [256,128], [256,256], and [1000,100]. We tested a range of L1 penalties on all of the weights of the network, from $10^{-7}$ to $10^{-2}$. All models attempted to optimize a least squares loss using the Adam optimizer, with learning rates again selected by hyperparameter tuning ranging from $10^{-5}$ to $10^{-3}$. Finally, we implemented an early stopping parameter of $20$ rounds to select the number of epochs of training (training was stopped after no improvement on validation error for 20 epochs, and the number of epochs was chosen based on optimal validation set error). We found that the optimal architecture (chosen by lowest validation set error) had two hidden layers of size 512 and 256, an L1 penalty on the weights of $10^{-3}$ and a learning rate of $10^{-5}$. We additionally found that 120 was the optimal number of training epochs.

\textbf{Attribution prior neural networks.} We next applied our attribution prior to the neural networks. First, we tuned networks to the optimal conditions described above. We then added extra epochs of fine-tuning where we ran an alternating minimization of the following objectives:

\begin{equation}
    \mathcal{L}(\theta;X,y) = \|f_\theta(X) - y \|_2^2 + \lambda \| \theta \|_1
\end{equation}
\begin{equation}
    \mathcal{L}(\theta;X) = \Omega_{graph}(\Phi(\theta,X)) = \nu \bar{\phi}^T L_G \bar{\phi}
\end{equation}

Following \cite{ross2017right}, we selected $\nu$ to be $100$ so that the $\Omega_{graph}$ term would initially be equal in magnitude to the least squares and L1 loss terms. We found that $5$ extra epochs of tuning were optimal by validation set error. We drew $k = 10$ background samples for our attributions. To test our attribution prior using gradients as the feature attribution method (rather than expected gradients), we followed the exact same procedure, only we replaced $\bar{\phi}$ with the average magnitude of the gradients rather than the expected gradients.

\textbf{Graph convolutional networks.} We followed the implementation of graph convolution described in \cite{DBLP:journals/corr/KipfW16}. The architectures were searched as follows: in every network we first had a single graph convolutional layer (we were limited to one graph convolution layer due to memory constraints on each Nvidia GTX 1080-Ti GPU that we used), followed by two fully connected layers of sizes (512,256), (512,128), or (256,128). We tuned over a wide range of hyperparameters, including L2 penalties on the weights ranging from $10^{-5}$ to $10^{-2}$, L1 penalties on the weights ranging from $10^{-5}$ to $10^{-2}$, learning rates of $10^{-5}$ to $10^{-3}$, and dropout rates ranging from $0.2$ to $0.8$. We found the optimal hyperparameters based on validation set error were two hidden layers of size 512 and size 256, an L2 penalty on the weights of $10^{-5}$, a learning rate of $10^{-5}$, and a dropout rate of $0.6$. We again used an early stopping parameter and found that 47 epochs was the optimal number.

\section{Sparsity experiments}
\label{sec:sparsityExperiments}
\subsection{Data description and processing}
Our sparsity experiments used data from the NHANES I survey \cite{miller1973plan} and contained 35 variables (expanded to 118 features by one-hot encoding of categorical variables) gathered from 13,000 patients. The measurements included demographic information like age, sex, and BMI as well as physiological measurements like blood, urine, and vital sign measurements. The prediction task was a binary classification of whether the patient was still alive (1) or not (0) 10 years after data were gathered. 

Data were mean-imputed and standardized so that each feature had 0 mean and unit variance. 
For each of the 200 experimental replicates, 100 train and 100 validation points were sampled uniformly at random; all other points were allocated to the test set.

\subsection{Model}
We trained a range of neural networks to predict survival in the NHANES data. The architecture, nonlinearities, and training rounds were all held constant at values that performed well on an unregularized network, and the type and degree of regularization were varied. All models used ReLU activations and a single output with binary cross-entropy loss; in addition, all models ran for 100 epochs with an SGD optimizer with learning rate 0.001 on the size-100 training data. The entire 100-sample training set fit in one batch. Because the training set was so small, all of its 100 samples were used for expected gradients attributions during training and evaluation, yielding $k=100$. Each model was trained on a single GPU on a desktop workstation with 4 Nvidia 1080 Ti GPUs.

\textbf{Architecture. }We considered a range of architectures, including single-hidden-layer 32-node, 128-node, and 512-node networks, two-layer [128,32] and [512,128]-node networks, and a three-layer [512,128,32]-node network; we fixed the [512,128,32] architecture for future experiments.

\textbf{Regularizers. }We tested a large array of regularizers in addition to those considered in the maintext. For details, see  Supplement Section \ref{sec:sparsemodel_supp}.

\subsection{Hyperparameter tuning}
\label{sec:methodtuning} We selected the hyperparameters for our models based on validation performance. We searched all L1, L2, SGL and attribution prior penalties with 121 points sampled on a log scale over $[10^{-7},10^5]$ (Supplementary Figure \ref{fig:supptune}). Other penalties, not displayed in the maintext experiments, are discussed in Supplement Section \ref{sec:sparseparams}.

\subsection{Maintext methods}

\hspace{10pt} \textbf{Performance and sparsity bar plots. }The performance bar graph (Figure \ref{fig:nhanes_reg}, top left) was generated by plotting mean test ROC-AUC of the best model of each type (chosen by validation ROC-AUC) averaged over each of the 200 subsampled datasets, with confidence intervals given by 2 times the standard error over the 200 replicates. The sparsity bar graph (Figure \ref{fig:nhanes_reg}, bottom left) was constructed using the same process, but with Gini coefficients rather than ROC-AUCs.

\textbf{Feature importance distribution plot. }The distribution of feature importances was plotted in the main text as a Lorenz curve (Figure \ref{fig:nhanes_reg}, bottom right): for each model, the features were sorted by global attribution value $\bar\phi_i$, and the cumulative normalized value of the lowest $q$ features was plotted, from 0 at $q=0$ to 1 at $q=p$. A lower area under the curve indicates more features had relatively small attribution values, indicating the model was sparser. Because 200 replicates were run on small subsampled datasets, the Lorenz curve for each model was plotted using the averaged mean absolute sorted feature importances over all replicates. Thus, for a given model type, the $q=1$ point represented the mean absolute feature importance of the least important feature averaged over each replicate, $q=2$ added the mean importance for the second least important feature averaged over each replicate, and so on.

\textbf{Performance vs sparsity plot. }Validation ROC-AUC and model sparsity were calculated for each of the 121 regularization strengths and averaged over each of the 200 replicates. These were plotted on a scatterplot to show the possible range of model sparsities and ROC-AUC performances (Figure \ref{fig:nhanes_reg}, top right) as well as the tradeoff between sparsity and performance.

\textbf{Statistical significance. }Statistical significance of the sparse attribution prior performance was assessed by comparing the test ROC-AUCs of the sparse attribution prior models on each of the 200 subsampled datasets to those of the other models (L1 gradients, L1 weights, SGL, and unregularized). Significance was assessed by 2-sided paired-samples  $T$-test, paired by subsampled dataset. The same process was used to calculate the significance of model sparsity as measured by the Gini coefficient. Detailed tables of the resulting $p$-values and test statistics $T$ are shown in Supplement Section \ref{sec:additionalmaintext}.

\subsubsection*{Code Availability}
Implementations of attribution priors for Tensorflow and PyTorch are available at \url{https://github.com/suinleelab/attributionpriors}. This repository also contains code reproducing main results from the paper.

\subsubsection*{Data Availability}
The data for all experiments and figures in the paper are publicly available. The repository above contains a downloadable version of the dataset used for the sparsity experiment, as well as links to download the datasets used in the image and graph prior experiments. Data for the benchmarks was published as part of \cite{lundberg2019explainable} and can be accessed at \url{https://github.com/suinleelab/treeexplainer-study/tree/master/benchmark}


\subsubsection*{Competing Interests}
The authors declare no competing interests.

\subsubsection*{Acknowledgments}
The results published here are partially based upon data generated by the Cancer Target Discovery and Development (CTD2) Network (https://ocg.cancer.gov/programs/ctd2/data-portal) established by the National Cancer Institute’s Office of Cancer Genomics.


\clearpage
\normalem
\printbibliography
\makeatletter\@input{genmainaux.tex}\makeatother
\end{document}


\appendix
\onecolumn
\section*{Supplementary Material}
\section{Related Work} \label{sec: supp_related_work}
There have been many previous attribution methods proposed for deep learning models \cite{lundberg2017unified, binder2016layer, shrikumar2017learning, sundararajan2017axiomatic}. We chose to extend integrated gradients because it is easy to differentiate and comes with theoretical guarantees.

Training with gradient penalties has also been discussed by existing literature. \cite{drucker1992improving} introduced the idea of regularizing the magnitude of model gradients in order to improve generalization performance on digit classification. Since then, gradient regularization has been used extensively as an adversarial defense mechanism in order to minimize changes to network outputs over small perturbations of the input \cite{jakubovitz2018improving, yu2018towards, roth2018adversarially}. \cite{ross2018improving} make a connection between gradient-based training for adversarial purposes and network interpretability.
\cite{ilyas2019notbugs} formally describe how the phenomena of adversarial examples may arise due to features that are predictive yet non-intuitive, and stress the need to incorporate human intuition into the training process.

Work on actually incorporating feature attribution methods into training is relatively recent. \cite{sen2018supervising} formally describe the problem of classifiers having unexpected behavior on inputs not seen in the training distribution, like those generated by asking whether a prediction would change if a particular feature value changed. They describe an active learning algorithm that updates a model based on points generated from a counter-factual distribution. Their work differs from ours in that they use feature attributions to generate counter-factual examples, but do not directly penalize the attributions themselves. \cite{ross2017right} introduce the idea of training models to have correct explanations, not just good performance. Their method is an instance of attribution priors in which the attribution function is gradients and the penalty function is minimizing the gradients of features known to be unimportant for each sample. \cite{chen2019robust, liu2019incorporating} both present attribution priors that penalize integrated gradients attributions. Our work improves on previous attribution priors in three ways. First, our broader interpretation of attribution priors allows us to instantiate three different penalty functions that encode human intuition without needing to know specific target values for each feature attribution. Second, our development of expected gradients provides a novel feature attribution method that can be regularized efficiently using a sampling procedure, allowing us to train with respect to more background references in a shorter time. Third, we empirically show that using an axiomatic method like expected gradients yields substantially better results than simple gradient-based priors. Overall, flexible attribution priors that use axiomatic feature attribution methods lead to more interpretable models with better performance than previous approaches.

\section{Comparison with Contextual decomposition explanation penalization}
\label{sec:supp_cdep}

In addition to the prior methods we compare our method with in the main text, we also compare to a contemporary method called contextual decomposition explanation penalization (CDEP) \cite{rieger2020interpretations}. This framework is closely related to attribution priors, with small changes:
\[ \theta  = \text{argmin}_\theta \mathcal{L}(\theta;X,y) + \lambda \sum_i \Omega_c(\Phi_c(\theta,x_i,s_i)) \]
where $\Phi_c$ now represents the attribution method contextual decomposition \cite{murdoch2018beyond} rather than our attribution method, expected gradients. Importantly, $\Phi_c$ produces an attribution for a single \textit{group} of features, rather than an attribution for each individual feature. We call the feature set whose attributions are desired for a given sample $s_i$. $\Omega_c$ takes the more limited form of $||\beta_i-\Phi_c(\theta,x_i,s_i)||_1$ where $\beta_i$ represents a target attribution for the feature set $s_i$.

Our approach offers substantial benefits when compared to CDEP. First, our method is significantly easier to use in practice. The attribution method Contextual Decomposition can not simply be applied to any architecture of network and requires custom decomposition rules for each type of layer.\footnote{As noted by the authors on the project's Github repository, ``the current CD implementation doesn't always work for all types of networks... you may need to write a custom function that iterates through the layers of your network.'' \cite{cdepgithub}} 
This means it may be difficult or impossible to use some architectures (i.e., attention-based networks) with Contextual Decomposition. 
Training attribution priors using expected gradients, on the other hand, does not require any knowledge of or adaptation to the specific network architecture in question. Any network with gradients that can be calculated by TensorFlow or PyTorch will automatically work with our approach.

Furthermore, we find that EG is more computationally efficient than CDEP and that the CDEP paper substantially underestimates the computational efficiency of our method. The paper compares CDEP, EG, and the Right for the Right Reasons (RRR) method \cite{ross2017right} on three datasets -- Grayscale Decoy, Color MNIST, and ISIC Skin Cancer. Results on the first two datasets are inconclusive: CDEP outperforms both EG and RRR on Color MNIST, while EG and RRR outperform it on Grayscale Decoy. For the third dataset, however, the authors only compare to RRR, claiming that EG is too slow and memory-intensive to run on the dataset.\footnote{ ``We were not able to compare against the method recently proposed in Erion et al., 2019 due to its prohibitively slow training and large memory requirements."\cite{rieger2020interpretations}}

In fact, we show that EG runs faster and with less memory usage than CDEP on the ISIC dataset. Using the code available in the CDEP authors' publicly-available repository \cite{cdepgithub}, we downloaded the data to reproduce their skin cancer data experiment. We evaluated the amount of memory and time required to train each batch of image data on the full VGG-16 network with each method. We were limited to batch sizes of 4 images for all methods because this was the largest batch size that could be run with CDEP on an NVIDIA GEFORCE RTX 2080 Ti GPU (EG and RRR both could accommodate larger batch sizes).

We found that while all three methods are slower than training without any kind of attribution prior, EG and RRR both train faster than CDEP (see Supplemental Figure \ref{fig:CDEPeval}). Furthermore, EG and RRR both require less GPU memory than CDEP. Finally, while the CDEP paper claims that EG requires substantially more memory and processing time than RRR, our results show that EG and RRR are in fact nearly identical with regard to speed and memory. 

\begin{figure}
    \centering
    \includegraphics[width=\linewidth]{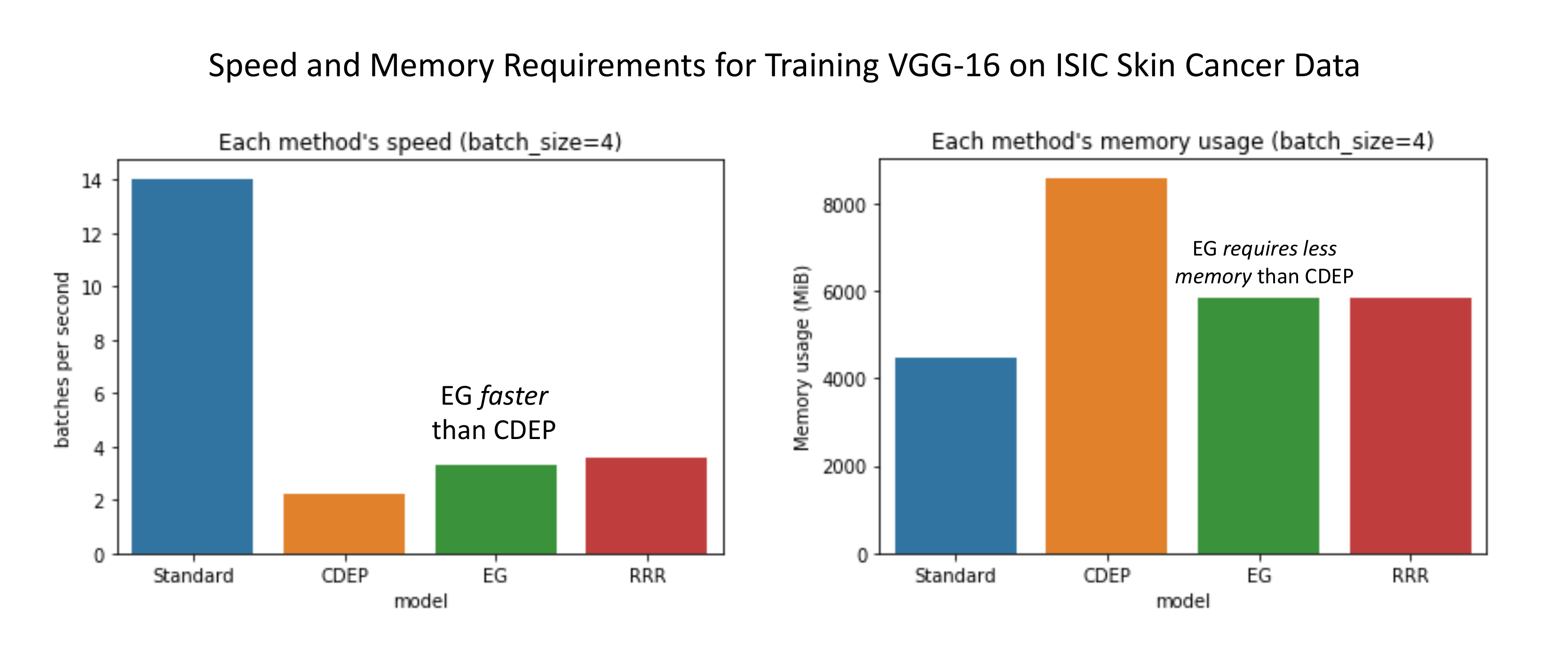}
    \caption{When training the full VGG-16 model, EG outperforms CDEP in terms of batches per second and memory requirement on ISIC Skin Cancer dataset.}
    \label{fig:CDEPeval}
\end{figure}

The authors of the CDEP paper are able to report substantially faster performance than we observe here because, in their experiments, they freeze the feature extracting layers of the VGG-16 (all layers except the final fully connected layers), and pre-compute CD attributions and image features for each image up to the end of the frozen layers in the network. When CDEP is run in identical conditions to EG and RRR, without these preprocessing steps, its performance is substantially worse. While the preprocessing steps used in CDEP may be useful ways to improve training speed, it is not accurate to conclude that contextual decomposition is a faster explanation method or that incorporating it into training as in CDEP is inherently more computationally efficient than EG -- in fact, we show the opposite is the case, and that the claimed speed benefits of CDEP are due to preprocessing steps rather than CDEP itself. 

It is also possible that the CDEP paper overestimates the resource requirements of EG and attribution priors by choosing the wrong hyperparameter $k$ (number of references). As we show in Sections 2.2 and 2.3, $k=1$ is sufficient to train high quality models in our image experiments. A setting much higher than $k=1$ could dramatically slow down EG without improving performance.

Finally, we observe that CDEP would be completely unsuitable as an attribution method for the types of priors we proposed in our work. Each prior suggested in our work requires attributions {for each individual feature}, and using EG we can attain those attributions during training at the cost of a single backward pass through the original model and first-order gradients with respect to each input.
To attain individual feature-level attributions using Contextual Decomposition would require an additional $d$ forward passes {for each batch during training} (where $d$ is the number of input features), as well as a backward pass through each of the $d$ resulting attributions, and thus would be entirely computationally infeasible. 

The CDEP framework is very efficient when attributions are desired for a \textit{small number of groups of features}. While it is completely impractical to calculate during training if \textit{many individual feature attributions} are desired, CDEP will likely outperform EG in terms of efficiency in the case when several conditions are met: 1) attributions are only necessary for a small number of groups of features, 2) only part of the network needs to be tuned, and 3) the network is of an architecture that CD can be applied to.

In conclusion, attribution priors offer several potential advantages over CDEP, including compatibility with both a greater variety of models and a greater variety of priors. In addition, the strongest criticisms leveled at attribution priors by the CDEP paper seem to stem from implementation error rather than real deficiencies in the method.

\section{Benchmarking Expected Gradients}
\label{sec:supp_benchmark}

\subsection{Sampling convergence}

\begin{figure}
    \centering
    \includegraphics[width=\linewidth]{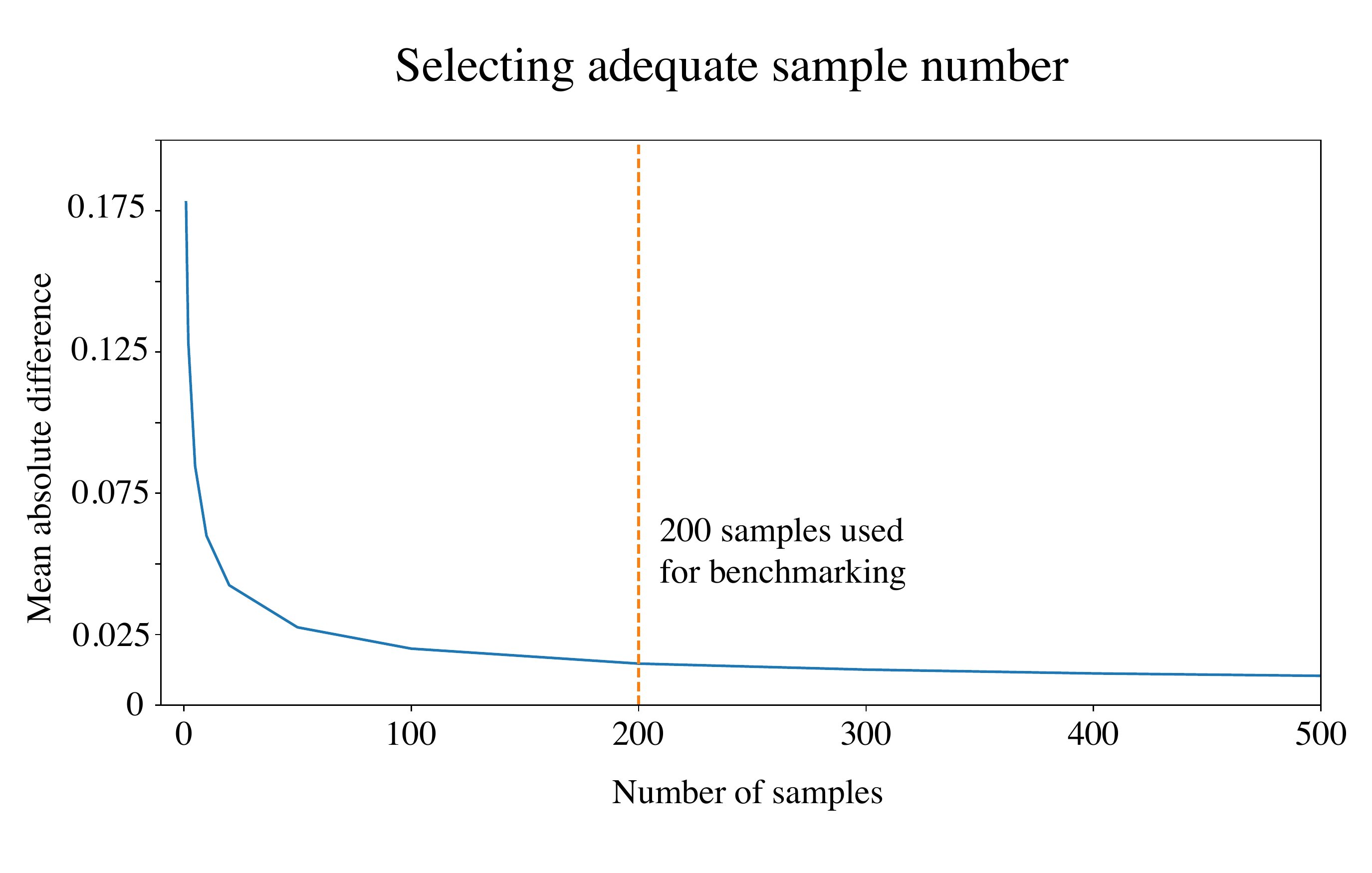}
    \caption{Feature attribution values attained using expected gradients converge as the number of background samples drawn is increased.}
    \label{fig:sampling_convergence}
\end{figure}

Since expected gradients reformulates feature attribution as an \textit{expected value} over two distributions (where background samples $x'$ are drawn from the data distribution and the linear interpolation parameter $\alpha$ is drawn from $U(0,1)$), we wanted to ensure that we are drawing an adequate number of background samples for convergence of our attributions when benchmarking the performance of our attribution method. Since our benchmarking code was run on the Correlated Groups 60 synthetic dataset, as a baseline we explain all $1000$ samples of this data sampling the full dataset (1000 samples) as background samples. To assess convergence to the attributions attained at this number of samples, we measure the mean absolute difference between the attribution matrices resulting from different numbers of background samples (see Figure~\ref{fig:sampling_convergence}). We empirically find that our attributions are well-converged by the time 100-200 background samples are drawn. Therefore, for the rest of our benchmarking experiments, we used 200 as the number of background samples. During training, even using the lowest possible setting of $k=1$, we end up drawing far more than 200 background samples over the course of an epoch (order of magnitude in the tens of thousands, rather than hundreds).

\subsection{Benchmark evaluation metrics}
\label{sec:supp_metrics}

To compare the performance of expected gradients with other feature attribution methods, we used the benchmark metrics proposed in \cite{lundberglocal}. These metrics were selected as they capture a variety of recent approaches to quantitatively evaluating feature importance estimates. For example, the Keep Positive Mask metric (KPM) is used to test how well an attribution method can find the features that lead to the greatest increase in the model's output. This metric progressively removes features by masking with their mean value, in order from least positive impact on model output to most positive impact on model output, as ranked by the attribution method being evaluated. As more features are masked, the model's output is increased, creating a curve. The KPM metric measures the area under this curve (larger area corresponds to better attribution method). In addition to the KPM metric, 17 other similar metrics (e.g. Remove Absolute Resample, Keep Negative Impute, etc.) were used (see supplementary material of \cite{lundberglocal} for more details on benchmark metrics). For all of these metrics, a larger number corresponds to a better attribution method. In addition to finding that Expected Gradients outperforms all other attribution methods on nearly all metrics tested for the dataset shown in Table 1 in the main text (the synthetic Correlated Groups 60 dataset proposed in \cite{lundberglocal}), we also tested all 18 metrics on another dataset proposed in the same paper (Independent Linear 60) and find that Expected Gradients is chosen as the best method by all metrics in that case as well (see Table~\ref{ind_lin_60_table}). The Independent Linear 60 dataset is comprised of 60 features, where each feature is a 0-mean, unit variance gaussian random variable plus gaussian noise, and the label to predict is a linear function of these features. The Correlated Groups 60 dataset is essentially the same, but now certain groups of 3 features have $0.99$ correlation.

For attribution methods to compare, we considered expected gradients (as described in the main text), integrated gradients (as described in \cite{sundararajan2017axiomatic}), gradients, and random.

\definecolor{nicered}{HTML}{cc002f}
\begin{table}
  \caption{Benchmark on Independent Linear 60 dataset}
  \label{ind_lin_60_table}
  \centering
  \begin{tabular}{llllllllll}
    \toprule
    Attribution Method & KPM & KPR & KPI & KNM & KNR & KNI & KAM & KAR & KAI \\
    \midrule
    Expected Gradients & \textcolor{nicered}{\textbf{4.096}} & \textcolor{nicered}{\textbf{4.179}} & \textcolor{nicered}{\textbf{4.264}} &  \textcolor{nicered}{\textbf{4.014}} &  \textcolor{nicered}{\textbf{3.835}} &  \textcolor{nicered}{\textbf{4.153}} &  \textcolor{nicered}{\textbf{0.941}} &  \textcolor{nicered}{\textbf{0.946}} & 
    \textcolor{nicered}{\textbf{0.938}} \\
    Integrated Gradients &
    4.055 &
    4.112 & 
    4.176 & 
    3.949 & 
    3.753 & 
    4.070 & 
    \textcolor{nicered}{\textbf{0.941}} & 
    0.945 &  
    \textcolor{nicered}{\textbf{0.938}} \\
    Gradients & 0.044 
    & 0.107
    & 0.029
    & 0.155
    & -0.150
    & 0.172
    & 0.902
    & 0.905
    & 0.902\\
    Random & -0.152 
    & 0.102
    & -0.152
    & 0.111
    & -0.126
    & 0.060
    & 0.470
    & 0.482
    & 0.438\\
    \bottomrule
    \toprule
    Attribution Method & RPM & RPR & RPI & RNM & RNR & RNI & RAM & RAR & RAI \\
    \midrule
    Expected Gradients 
    & \textcolor{nicered}{\textbf{4.079}}
    & \textcolor{nicered}{\textbf{3.941}}
    & \textcolor{nicered}{\textbf{4.210}}
    & \textcolor{nicered}{\textbf{4.203}}
    & \textcolor{nicered}{\textbf{4.260}}
    & \textcolor{nicered}{\textbf{4.356}}
    & \textcolor{nicered}{\textbf{0.992}}
    & \textcolor{nicered}{\textbf{0.977}}
    & \textcolor{nicered}{\textbf{1.019}}\\
    Integrated Gradients 
    & 4.013
    & 3.854
    & 4.113
    & 4.157
    & 4.186
    & 4.259
    & 0.973
    & 0.966
    & 0.995\\
    Gradients 
    & 0.110
    & -0.125
    & 0.133
    & 0.057
    & 0.080
    & 0.041
    & 0.947
    & 0.936
    & 0.985\\
    Random 
    & 0.012
    & -0.124
    & 0.059
    & 0.035
    & 0.101
    & 0.070
    & 0.504
    & 0.521
    & 0.527\\
    \bottomrule
  \end{tabular}
\end{table}

\section{Expected Gradients on ImageNet}
\label{sec:supp_imagenet}

\begin{figure}
    \centering
    \includegraphics[width=1.0\linewidth]{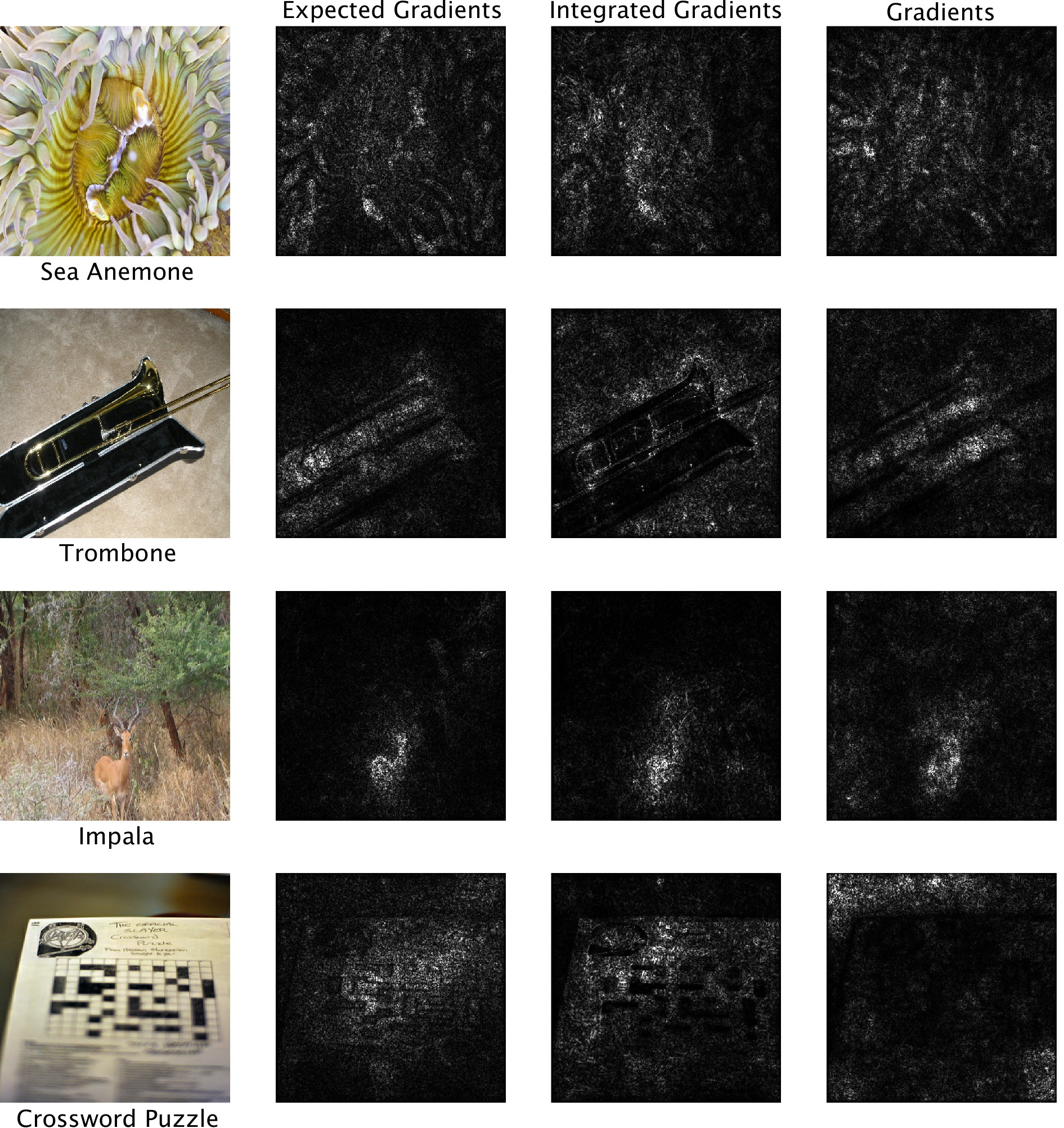}
    \caption{ A comparison of attribution methods on ImageNet. Integrated gradients fails to highlight black pixels as important when black is used as a baseline input. }
    \label{fig:eg_ig_black}
\end{figure}

One unfortunate consequence of choosing an arbitrary baseline point for methods like integrated gradients is that the baseline point by definition is unimportant. That is, if a user chooses the constant black image as the baseline input, then purely black pixels will not be highlighted as important by integrated gradients. This is true for any constant baseline input. Since expected gradients integrates over a dataset as its baseline input, it avoids forcing a particular pixel value to be unimportant. To demonstrate this, we use the inception v4 network trained on the ImageNet 2012 challenge \cite{szegedy2017inception, russakovsky2015imagenet}. We restore pre-trained weights from the Tensorflow Slim library \cite{silberman2016tensorflow}. In Figure \ref{fig:eg_ig_black}, we plot attribution maps of both expected gradients and integrated gradients as well as raw gradients. Here, we use the constant black image as a baseline input for integrated gradients. For both attribution methods, we use 200 sample/interpolation points. The figure demonstrates that integrated gradients fails to highlight black pixels.

\section{CIFAR-10 Experiments}
\label{sec:supp_cifar}

\subsection{Choosing Lambda}
\begin{figure}
    \centering
    \includegraphics[width=0.7\linewidth]{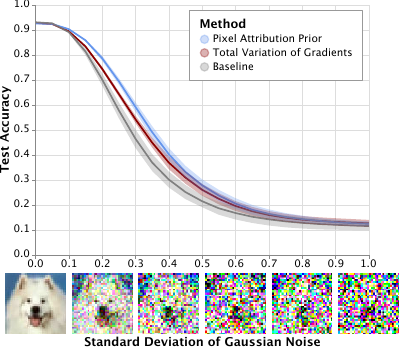}
      \caption{ Robustness to noise on CIFAR-10 with a stricter $\lambda$ threshold. Here, there is little difference in test accuracy on the original test set between the baseline and the image attribution prior model ($0.930 \pm 0.002$ for the baseline vs. $0.925 \pm 0.002$ for the pixel attribution prior). Both the image attribution prior model and the gradient-based model afford small improvements in robustness compared to the baseline. As in the main text, results here are the mean and standard deviation across 5 random initializations. }
    \label{fig:CIFAR10_small}
\end{figure}

In the main text, we demonstrated the robustness of the image attribution prior model with $\lambda$ chosen as the value that minimized the total variation of attributions while keeping test accuracy within $10\%$ of the baseline model. This corresponds to $\lambda = 0.001$ for both gradients and expected gradients if we search through 20 values logarithmically spaced in the range $[10^{-20}, 10^{-1}]$.
If instead, we choose the $\lambda$ that minimizes total variation of attributions while keeping test accuracy equivalent to the baseline model (within $1\%$), we see that both the attribution prior and regularizing the gradients provides modest robustness to noise. This corresponds to $\lambda = 0.0001$ for both gradients and expected gradients. We show this result in Figure \ref{fig:CIFAR10_small}.

\begin{figure}
        \centering
    \includegraphics[width=0.48\linewidth]{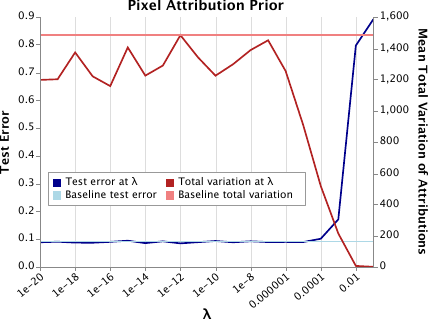}
    \includegraphics[width=0.48\linewidth]{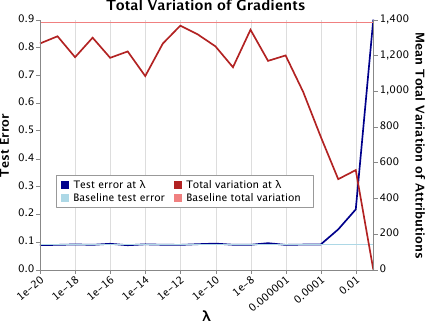}
    \caption{ Plotting the trade-off between accuracy and minmizing total variation of expected gradients (left) or gradients (right). For both methods, there is a clear elbow point after which test accuracy degrades to no better than random. The total variation of attributions is judged based on the attribution being penalized: expected gradients for the left plot, gradients for the right plot. }
    \label{fig:CIFAR10_tradeoff}
\end{figure}

For both the gradient-based model and the image attribution prior model, we also plot test accuracy and total variation of the attributions (gradients or expected gradients, respectively) in Figure \ref{fig:CIFAR10_tradeoff}. The $\lambda$ values we use correspond to the immediate two values before test accuracy on the original test set breaks down entirely for both the gradient and image attribution prior model. 

\section{MNIST Experiments}
\label{sec:supp_mnist}

\subsection{Results}
We choose $\lambda$ by sweeping over values in the range $[10^{-20}, 10^{-1}]$. We choose the $\lambda$ that minimizes the total variation of attributions such that the test error is within $1\%$ of the test error of the baseline model, which corresponds to $\lambda = 0.01$ for both the gradient model and the pixel attribution prior model. In Figure \ref{fig:mnist_results}, we plot the robustness of the baseline, the model trained with an attribution prior, and the model trained by penalizing the total variation of gradients. We find that on MNIST, penalizing the gradients does similarly to penalizing expected gradients. We also find that it is easier to achieve high test set accuracy and robustness simultaneously. 

\subsection{Attribution Maps}
\begin{figure}
        \centering
    \includegraphics[width=0.3\linewidth]{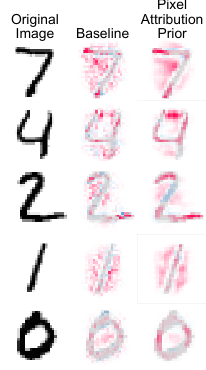}
    \hspace{0.2cm}
    \includegraphics[width=0.6\linewidth]{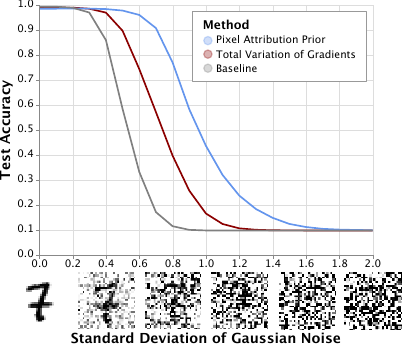}
    \caption{ Left: Expected gradients attributions (with 100 samples) on MNIST for both the baseline model and the model trained with an attribution prior, for five randomly selected images classified correctly by both the baseline and the regularized model. Red pixels indicate pixels positively influencing the prediction, while blue pixels negatively influence the prediction. Training with an attribution prior generates visually smoother attribution maps and tend to better highlight relevant parts of the image. Right: Training with an attribution prior induces robustness to noise, more so than an equivalent model trained by minimizing the total variation of gradients or an equivalent baseline. The baseline model achieves an accuracy of 0.9925, compared 0.9836 for the pixel attribution prior and 0.9888 for the gradient model. }
    \label{fig:mnist_results}
\end{figure}

In Figure \ref{fig:mnist_results} we plot the attribution maps of the baseline model compared to the model regularized with an image attribution prior. We find that the model trained with an image attribution prior more smoothly highlights the digit in the image.

\section{ImageNet Experiments}
\label{sec:ImageNetExperiments}

In this section, we detail experiments performed on applying $\Omega_{\textrm{pixel}}$ to classifiers trained on the ImageNet 2012 challenge \cite{russakovsky2015imagenet}. 

\subsection{Experimental Setup}

We use the VGG16 architecture introduced by \cite{simonyan2014very}. We then fine-tune the network from its pre-trained weights in the Tensorflow Slim package \cite{silberman2016tensorflow}. We fine-tune on the ImageNet 2012 training set using the original cross entropy loss function in addition to $\Omega_{\textrm{pixel}}$ using asynchronous gradient updates with a batch size of 16 split across 4 Nvidia 1080 Ti GPUs. During fine-tuning, we use the same training procedure outlined by \cite{silberman2016tensorflow}. This includes randomly cropping training images to $224 \times 224$ pixels, randomly flipping images horizontally, and normalizing each image to the same range. To optimize, we use gradient descent with a learning rate of $0.00001$ and a momentum of $0.9$. We use a weight decay of $0.0005$, and set $\lambda = 0.00001$ for the first epoch of fine-tuning, and $\lambda = 0.00002$ for the second epoch of fine-tuning. As with the MNIST experiments, we normalize the feature attributions before taking total variation. 

\subsection{Results}
\begin{figure}
    \centering
    \includegraphics[width=0.9\linewidth]{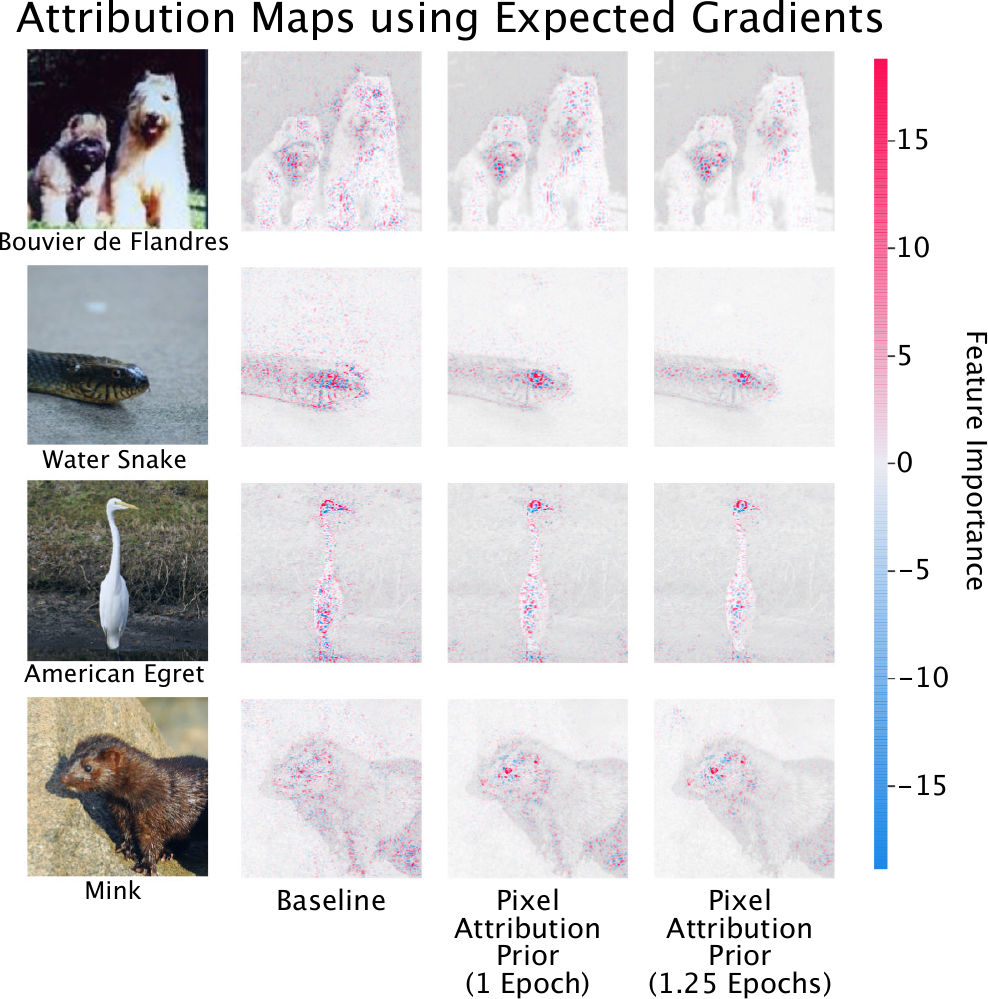}
    \caption{ Attribution maps generated by Expected Gradients on the VGG16 architecture before and after fine-tuning using an attribution prior. }
    \label{fig:EG_Imagenet}
\end{figure}

\begin{figure}
    \centering
    \includegraphics[width=0.9\linewidth]{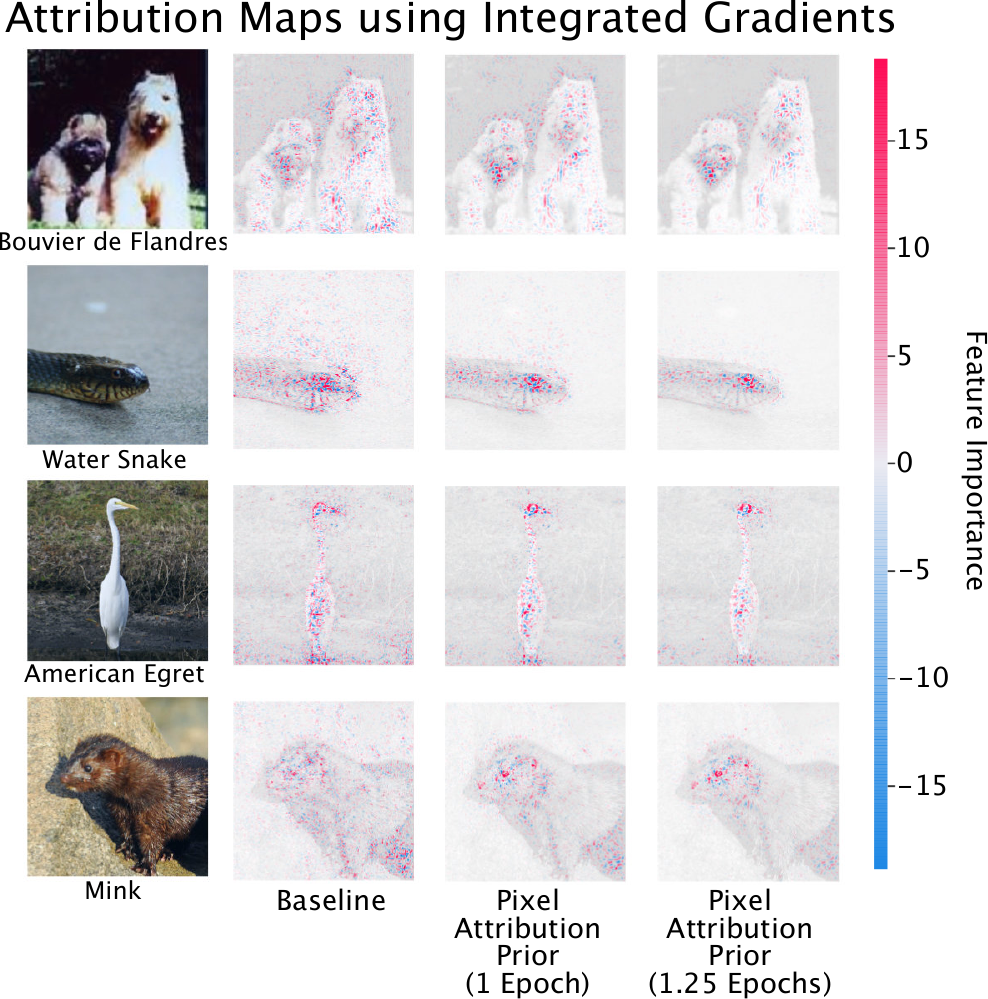}
    \caption{ Attribution maps generated by Integrated Gradients on the VGG16 architecture before and after fine-tuning using an attribution prior. }
    \label{fig:IG_Imagenet}
\end{figure}

\begin{figure}
    \centering
    \includegraphics[width=0.9\linewidth]{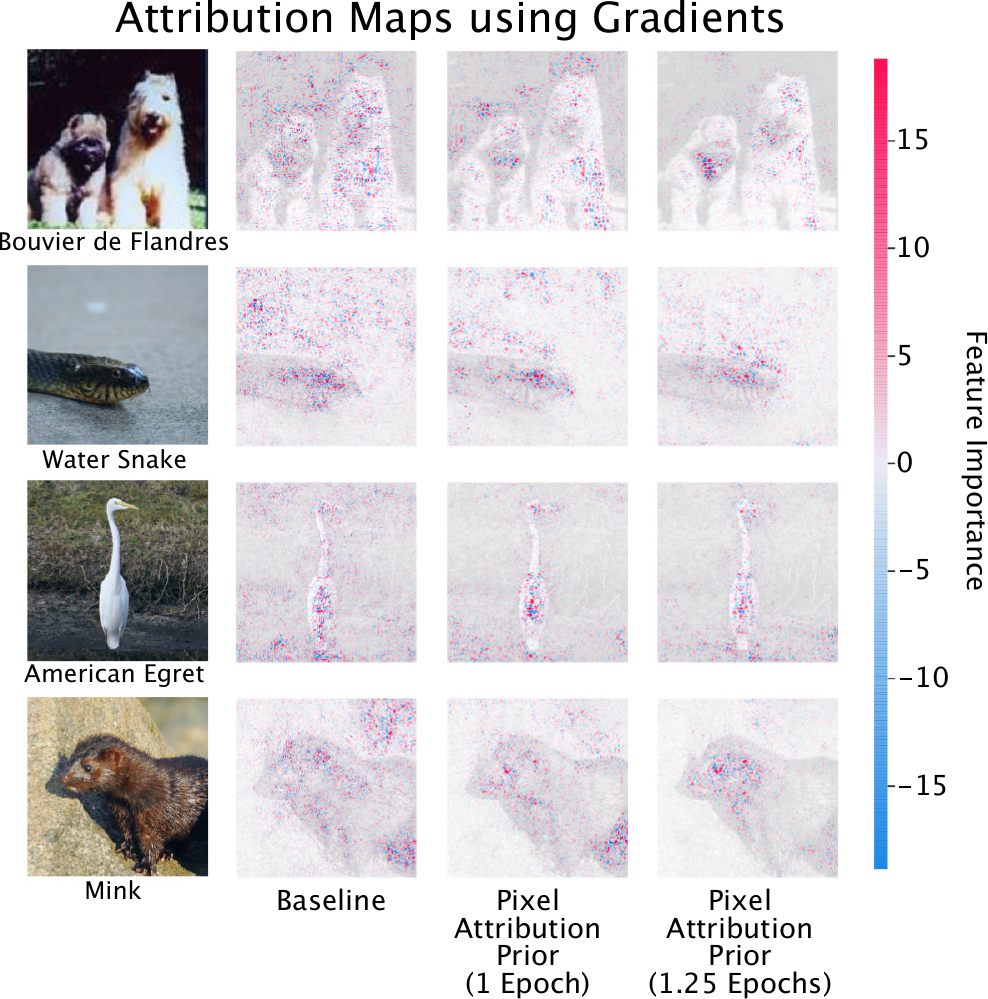}
    \caption{ Attribution maps generated by raw gradients on the VGG16 architecture before and after fine-tuning using an attribution prior. }
    \label{fig:Grad_Imagenet}
\end{figure}

We plot the attribution maps on images from the validation set using expected gradients for the original VGG16 weights (Baseline), as well as fine-tuned for 320,292 steps (Image Attribution Prior 1 Epoch) and fine-tuned for 382,951 steps, in which the last 60,000 steps were with twice the $\lambda$ penalty (Image Attribution Prior 1.25 Epochs). Figure \ref{fig:EG_Imagenet} demonstrates that fine-tuning using our penalty results in sharper and more interpretable image maps than the baseline network. In addition, we also plot the attribution maps generated by two other methods: integrated gradients (Figure \ref{fig:IG_Imagenet}) and raw gradients (Figure \ref{fig:Grad_Imagenet}). Networks regularized with our attribution prior show more clear attribution maps using any of the above methods, which implies that the network is actually viewing pixels more smoothly, independent of the attribution method chosen.

We note that in practice, we observe similar trade-offs between test accuracy and interpretability/robustness mentioned in \cite{ilyas2019notbugs}. We show the validation performance of the VGG16 network before and after fine-tuning in Table \ref{Tab:Imagenet_perf} and observe that the validation accuracy does decrease. 

\begin{table}
  \caption{ Performance of the VGG16 architecture on the ImageNet 2012 validation dataset before and after fine-tuning. }
  \label{Tab:Imagenet_perf}
  \centering
  \begin{tabular}{lll}
    \toprule
    Model & Top 1 Accuracy & Top 5 Accuracy \\
    \midrule
    Baseline & 0.709 & 0.898 \\
    Image Attribution Prior 1 Epoch & 0.699 & 0.886 \\
    Image Attribution Prior 1.25 Epochs & 0.674 & 0.876
    \end{tabular}
\end{table}

\section{Biological experiments}
\label{sec:supp_biological}

\subsection{RNA-seq preprocessing}

To ensure a quality signal for prediction while removing noise and batch effects, it is necessary to carefully preprocess RNA-seq gene expression data. For the biological data experiments, RNA-seq were preprocessed as follows:

\begin{enumerate}
    \item First, raw transcript counts were converted to fragments per kilobase of exon model per million mapped reads (FPKM). FPKM is more reflective of the molar amount of a transcript in the original sample than raw counts, as it normalizes the counts for different RNA lengths and for the total number of reads \cite{Mortazavi2008}. FPKM is calculated as follows:
    \begin{equation}
        FPKM = \frac{X_i \times 10^9}{Nl_i}
    \end{equation}
    Where $X_i$ is the raw counts for a transcript, $l_i$ is the effective length of the transcript, and $N$ is the total number of counts.
    \item Next, we removed non-protein-coding transcripts from the dataset.
    \item We removed transcripts that were not meaningfully observed in our dataset by dropping any transcript where $>70\%$ measurements across all samples were equal to 0.
    \item We $\log_2$ transformed the data
    \item We standardized each transcript across all samples, such that the mean for the transcript was equal to zero and the variance of the transcript was equal to one:
    \begin{equation}
        X_i^{'} = \frac{X_i - \mu_i}{\sigma_i}
    \end{equation}
    where $X_i$ is the expression for a transcript, $\mu_i$ is the mean expression of that transcript, and $\sigma_i$ is the standard deviation of that transcript across all samples.
    \item Finally, we corrected for batch effects in the measurements using the ComBat tool available in the sva R package \cite{10.1371/journal.pgen.0030161}.
\end{enumerate}

\subsection{Further details on experimental results}

\begin{figure}
    \centering
    \includegraphics[width=0.75\linewidth]{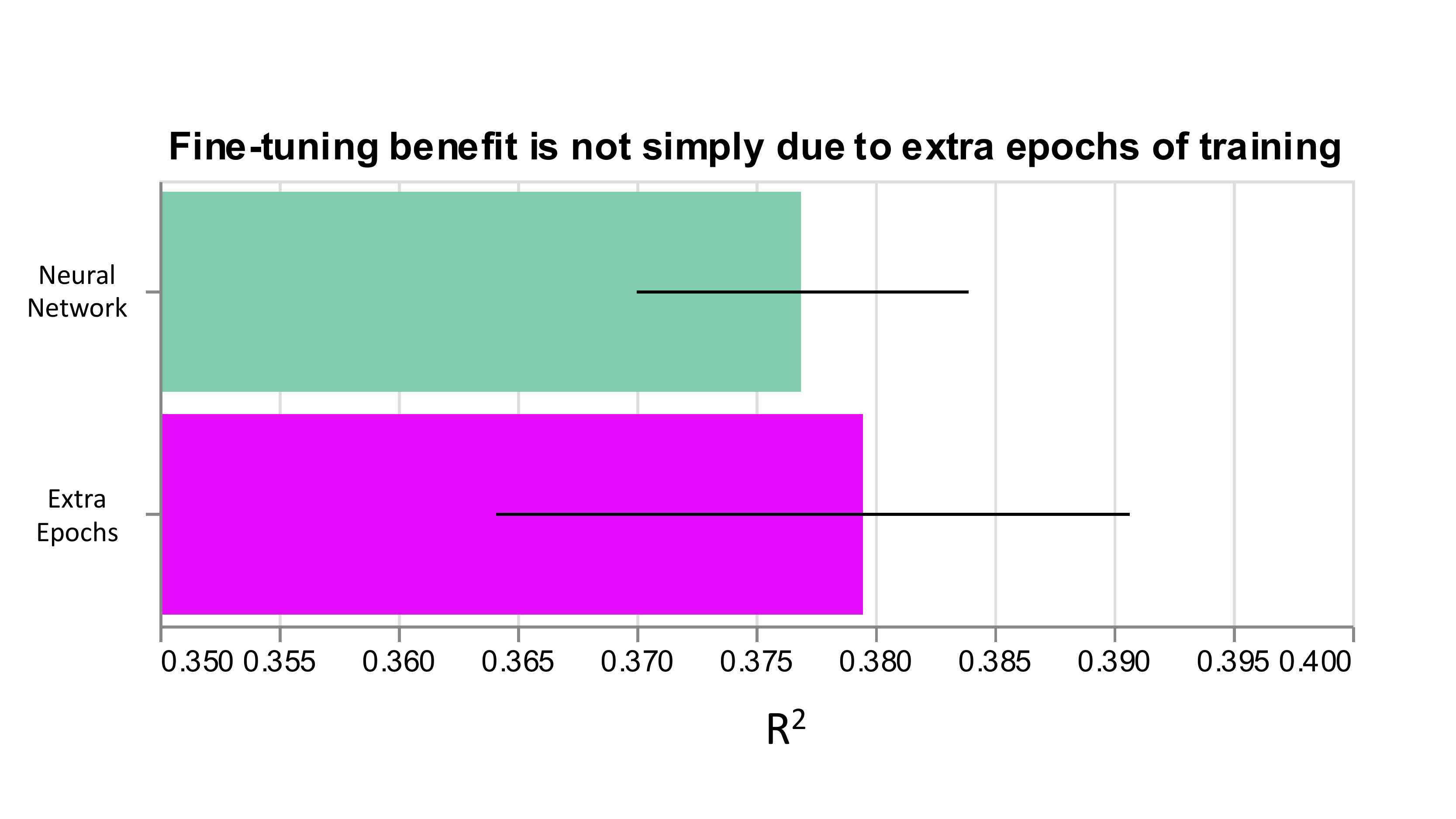}
    \caption{ Fine tuning \textit{without} graph prior penalty leads to no significant improvement in model performance. }
    \label{fig:ex_ep}
\end{figure}

Since we added graph-regularization to our model by fine-tuning, we wanted to ensure that the improved performance did not simply come from the additional epochs of training \textit{without} the attribution prior. We use a $t$-test to compare the $R^2$ attained from 10 independent retrainings of the regular neural network to the $R^2$ attained from 10 independent retrainings of the neural network with the same number of additional epochs that were optimal when adding the graph penalty (see Figure~\ref{fig:ex_ep}). We found no significant difference between the test error of these models ($p = 0.7565$).

\begin{figure}
    \centering
    \includegraphics[width=0.90\linewidth]{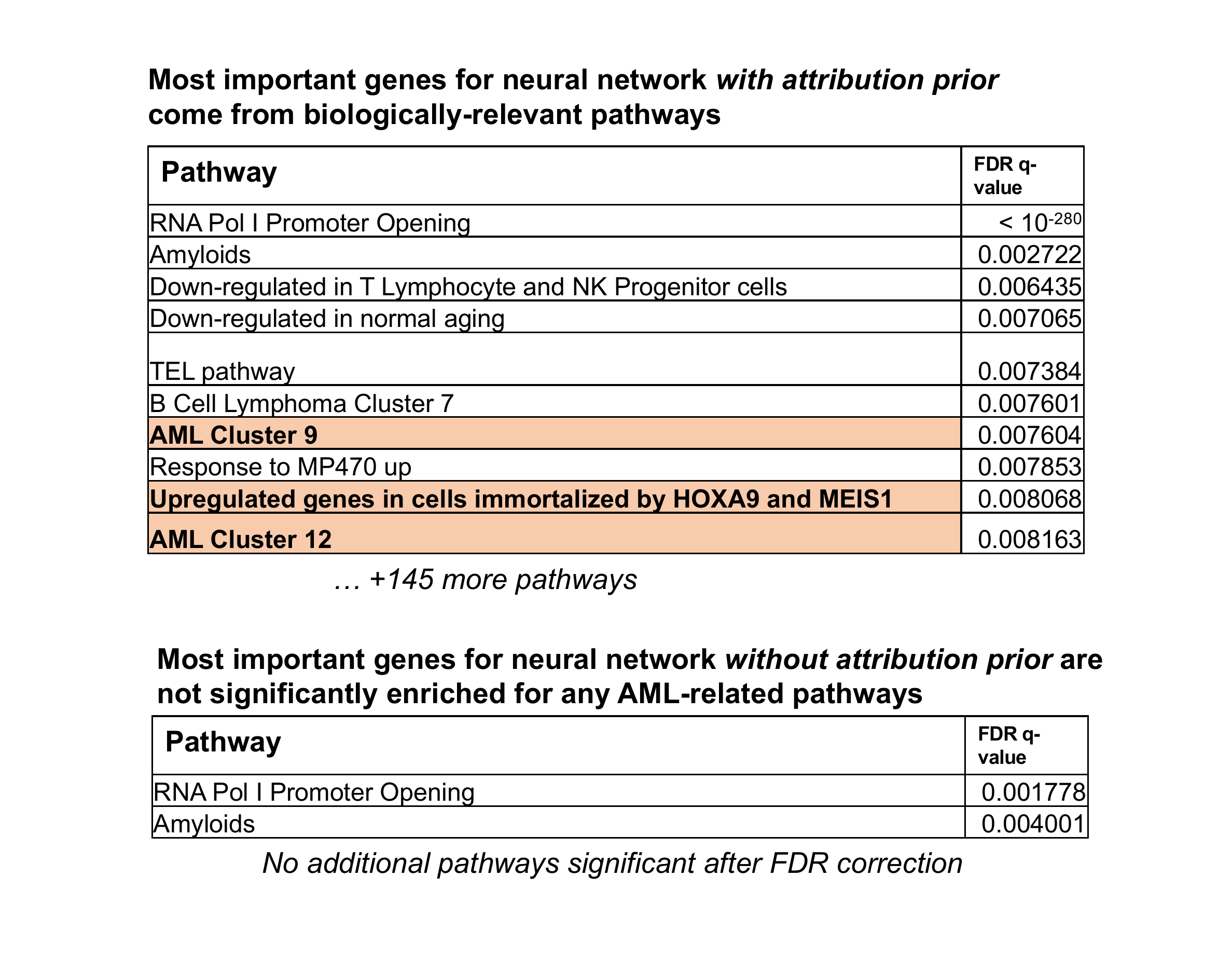}
    \caption{ Top pathways for neural networks with and without attribution priors }
    \label{fig:top_path}
\end{figure}

\begin{figure}
    \centering
    \includegraphics[width=0.90\linewidth]{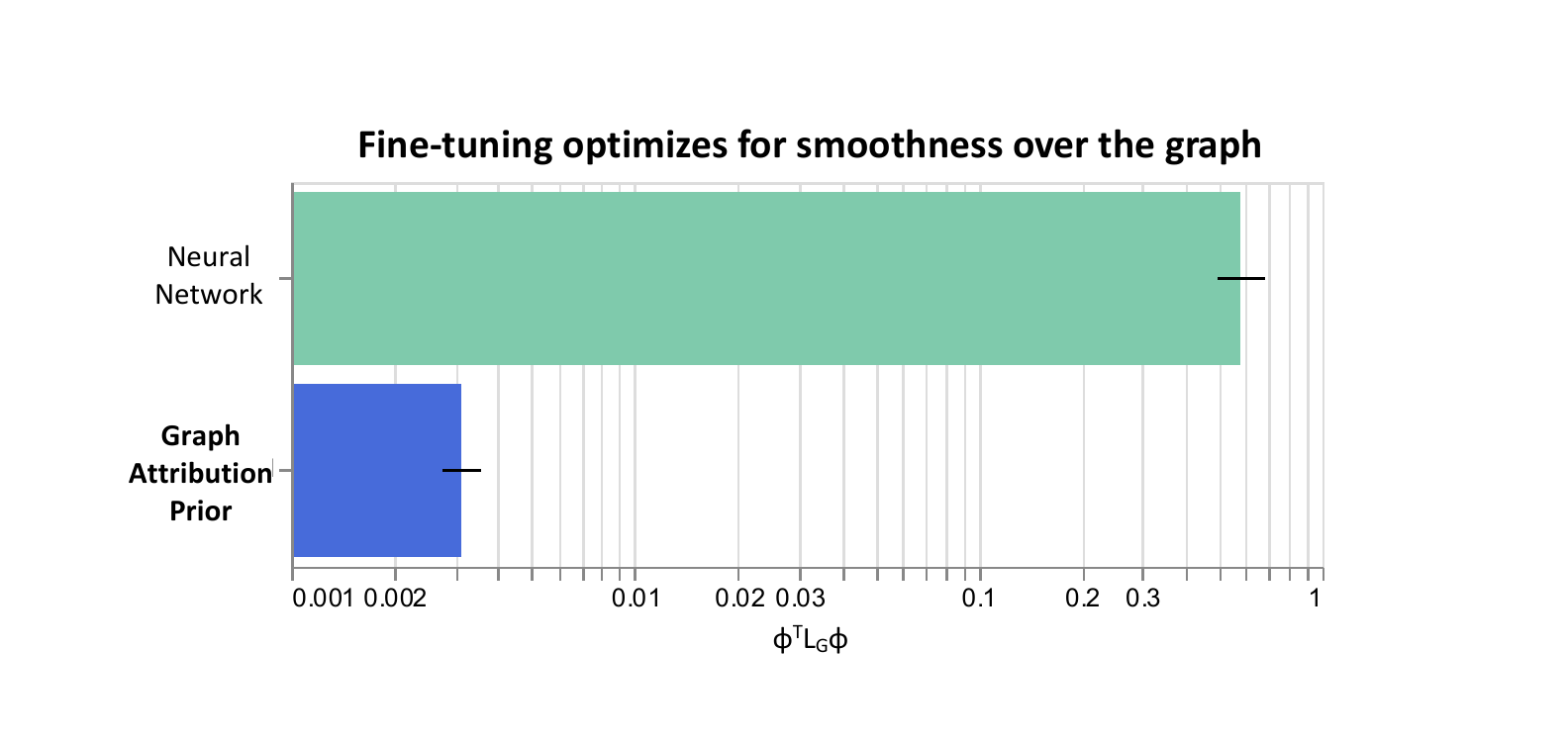}
    \caption{ Fine tuning optimizes for the metric we care about: smoothness over the graph }
    \label{fig:graph_smooth}
\end{figure}

To ensure that the models were learning the attribution metric we tried to optimize for, we compared the explanation graph penalty ($\bar{\phi}^T L_G \bar{\phi}$) between the unregularized and regularized models, and found that the graph penalty was on average nearly two orders of magnitude lower in the regularized models (see Figure~\ref{fig:graph_smooth}). We also examined the pathways that our top attributed genes were enriched for using Gene Set Enrichment Analysis and found that not only did our graph attribution prior model capture far more significant pathways, it also captured far more AML-relevant pathways (see Figure~\ref{fig:top_path}). We defined AML-relevant by a query for the term ``AML,'' as well as queries for AML-relevant transcription factors.

\section{Sparsity experiments}
\label{sec:supp_sparsity}

\subsection{Model}
\label{sec:sparsemodel_supp}

\textbf{Regularizers: }We tested a large array of regularizers. See the Methods section and section \ref{sec:sparseparams} for details on how optimal regularization strength was found for each regularizer. \textit{Italicized} entries were displayed in the main text. The optimal regularization strength from validation-set tuning is listed in the table below.
\begin{itemize}
\item \textit{Sparse Attribution Prior} - $\Omega_\textrm{sparse}$ as defined in the main text. 
\item Mixed L1/Sparse Attribution Prior - Motivated by the observation that the Gini coefficient is normalized and only penalizes the \textit{relative distribution} of global feature importances, we attempted adding an L1 penalty to ensure the attributions also remain small in an absolute sense. The L1 and Gini terms were equally weighted. This did not result in statistically significant improvements to performance or sparsity (subsection \ref{sec:additionalmaintext}).
\item \textit{Sparse Group Lasso} - Rather than simply encouraging the weights of the first-layer matrix to be zero, the sparse group lasso also encourages entire columns of the matrix to shrink together by placing an L2 penalty on each column. As in \cite{scardapane2017group}, we added a weighted sum of column-wise L2 norms to the L1 norms of each layer's matrix, without tuning the relative contribution of the two norms (equal weight on both terms). We also follow \cite{scardapane2017group} and penalize the absolute value of the \textit{biases} of each layer as well. 
\item \textit{Sparse Group Lasso First Layer }- This penalty was similar to \cite{scardapane2017group}, but instead of penalizing \textit{all} weights and biases, only the first-layer weight matrix was penalized. This model outperformed the SGL implementation adapted from \cite{scardapane2017group}, but did not outperform the sparse attribution prior. 
\item L1 First-Layer - In order to facilitate sparsity, we placed an L1 penalty on the input layer of the network. No regularization was placed on subsequent layers. 
\item \textit{L1 All Layers} - This penalty places an L1 penalty on all matrix multiplies in the network (not just the first layer). 
\item L1 Expected Gradients - This penalty penalizes the L1 norm of the vector of global feature attributions, $\bar\phi_i$ (analogous to how LASSO penalizes the weight vector in linear regression). 
\item L2 First-Layer - This penalty places an L2 penalty on the input layer of the network, with no regularization on subsequent layers. 
\item L2 All Layers - This penalty places an L2 penalty on all matrix multiplies in the network (not just the first layer). 
\item L2 Expected Gradients - This penalty penalizes the L2 norm of the vector of global feature attributions, $\bar\phi_i$ (analogous to how ridge regression penalizes the weight vector in linear models). 
\item Dropout - This penalty "drops out" a fraction $p$ of nodes during training, but uses all nodes at test time. 
\item \textit{Baseline (Unregularized)} - Our baseline model used no regularization. 
\item \textit{L1 Gradients} - To achieve the closest match to work by \cite{ross2017right,ross2017neural}, we placed a L1 penalty on the global gradients attribution vector of the network (mean across all samples of the absolute value of the gradient for each feature). This is similar to the "neural LASSO" of \cite{ross2017neural}, but with a goal of global sparsity (a model that uses few features overall) rather than local sparsity (a model that uses a small number of possibly different features for each sample). 
\item \textit{Gini Gradients} - An intermediate step between \cite{ross2017right,ross2017neural} and our sparse attribution prior would use gradients as an attribution, but our Gini coefficient-based sparsity metric as a penalty. In this model we encouraged a large Gini coefficient of the mean absolute value of the gradients attributions of the model, averaged over all samples. 
\end{itemize}
The maintext figures, with experiments repeated 200 times, compared the sparse attribution prior to  methods previously used in literature on sparsity in deep networks -- the L1 penalty on all layers, the sparse group lasso methods \cite{scardapane2017group}, and the L1 gradients penalty \cite{ross2017neural}. We also evaluated the Gini gradients penalty 
in these experiments. The other methods were evaluated similarly but not displayed in the maintext for space reasons and because there was less literature support. The optimal regularization parameters for the methods evaluated in the maintext are as follows (median over all replicates):

\begin{tabular}{l|r}
\toprule
{} &    Lambda \\
\midrule
Sparse Attribution Prior       & 1.259e+00 \\
Mixed L1/Sparse Attribution Prior    & 1.259e+00 \\
L1 (All Parameters)   & 3.162e+02 \\
L2 (All Parameters)   & 6.310e-01 \\
SGL (All Parameters)       & 1.000e-03 \\
SGL (First Layer)     & 3.981e+00 \\
L1 (First Layer)  & 1.000e+04 \\
L2 (First Layer) & 1.790e+03 \\
L1 (Expected Gradients)         & 5.012e+00 \\
L2 (Expected Gradients)         & 1.000e-02 \\
Dropout      & 6.598e-01 \\
L1 (Gradients)       & 1.995e+00 \\
Gini (Gradients)     & 1.259e+00 \\
Unregularized        & N/A \\
\bottomrule
\end{tabular}

\subsection{Hyperparameter tuning: }
\label{sec:sparseparams}
\begin{figure}
    \centering
    \includegraphics[width=\linewidth]{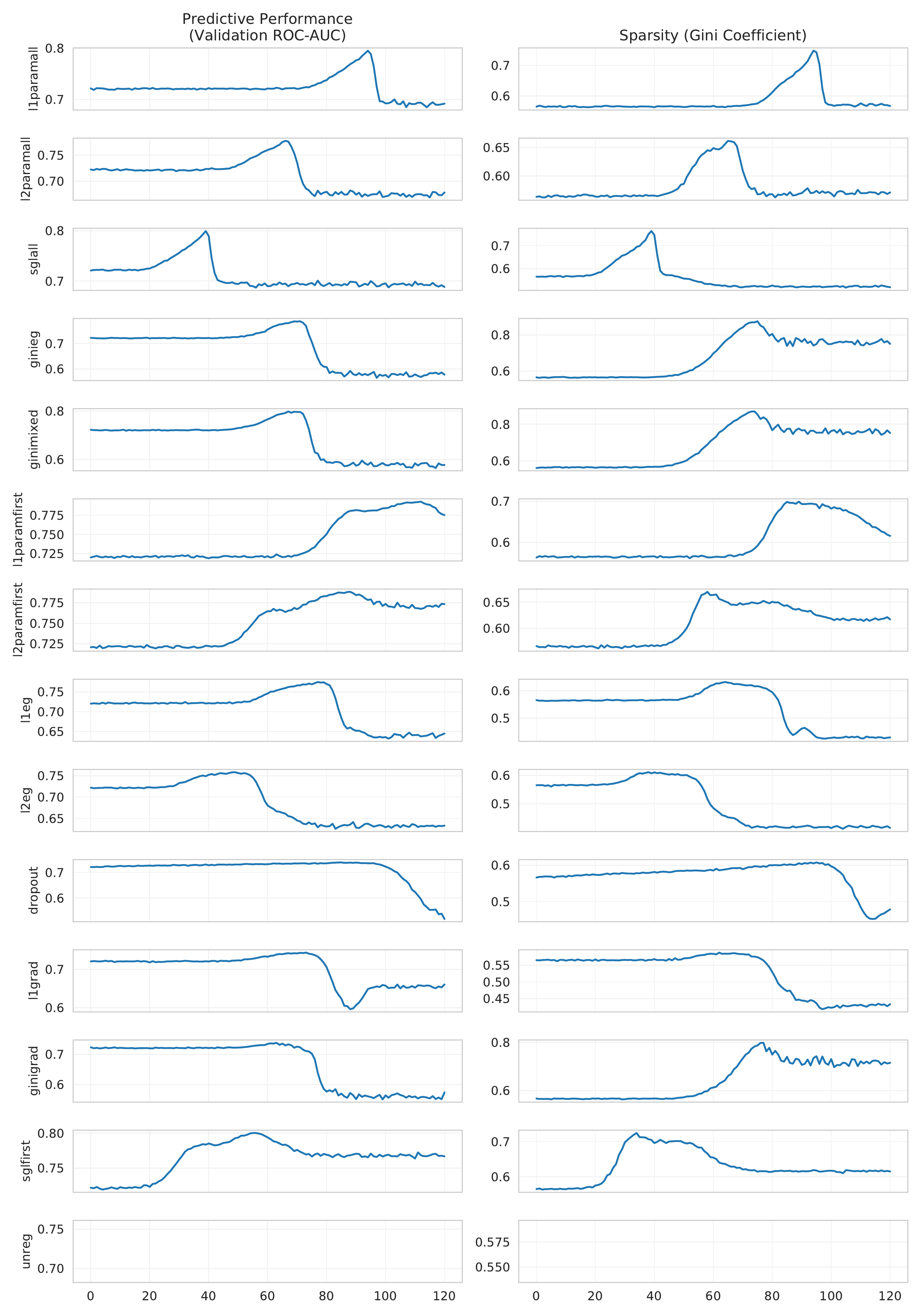}
    \caption{Validation performance and Gini coefficient as a function of regularization strength for all models, averaged over 200 subsampled datasets.}
    \label{fig:supptune}
\end{figure}

Most parameter tuning is discussed in the maintext and Methods sections. Dropout required special tuning; we tuned the dropout probability with 121 points linearly spaced over $(0,1]$.  Mean validation performance and sparsity across the range of tuning parameters is shown in Figure \ref{fig:supptune}.

\subsection{Additional Results (Maintext Experiments)}
\label{sec:additionalmaintext}
\textbf{Statistical significance: }Statistical significance was assessed as discussed in the Methods section. The resulting $p$-values and test statistics $T$ for the performance difference between the sparse attribution prior and the other methods were:

\begin{tabular}{l|rrrr}
\toprule
{} &  ROC-AUC (p) &  ROC-AUC (T) &  Sparsity (p) &  Sparsity (T) \\
\midrule
L1 (All Parameters)   &    2.523e-05 &    4.314e+00 &     1.188e-37 &     1.602e+01 \\
L2 (All Parameters)   &    1.251e-28 &    1.308e+01 &     4.477e-78 &     3.098e+01 \\
SGL (All Parameters)       &    1.400e-07 &    5.461e+00 &     1.540e-33 &     1.468e+01 \\
Mixed L1/Sparse Attribution Prior    &    7.697e-01 &   -2.932e-01 &     6.169e-01 &     5.010e-01 \\
L1 (First Layer) &    5.209e-07 &    5.188e+00 &     4.627e-65 &     2.567e+01 \\
L2 (First Layer) &    6.979e-17 &    9.148e+00 &     5.446e-88 &     3.548e+01 \\
L1 (Expected Gradients)         &    7.143e-24 &    1.152e+01 &     7.315e-85 &     3.401e+01 \\
L2 (Expected Gradients)          &    1.926e-39 &    1.661e+01 &     6.557e-88 &     3.544e+01 \\
Dropout      &    5.555e-41 &    1.712e+01 &    2.667e-100 &     4.170e+01 \\
L1 (Gradients)        &    1.459e-44 &    1.831e+01 &     3.923e-97 &     4.002e+01 \\
Gini (Gradients)     &    7.793e-35 &    1.510e+01 &     3.612e-27 &     1.261e+01 \\
SGL (First Layer)     &    4.571e-14 &    8.128e+00 &     3.667e-79 &     3.146e+01 \\
Unregularized        &    9.821e-56 &    2.220e+01 &    1.246e-107 &     4.579e+01 \\
\bottomrule
\end{tabular}

\begin{figure}
    \centering
    \includegraphics[width=0.75\linewidth]{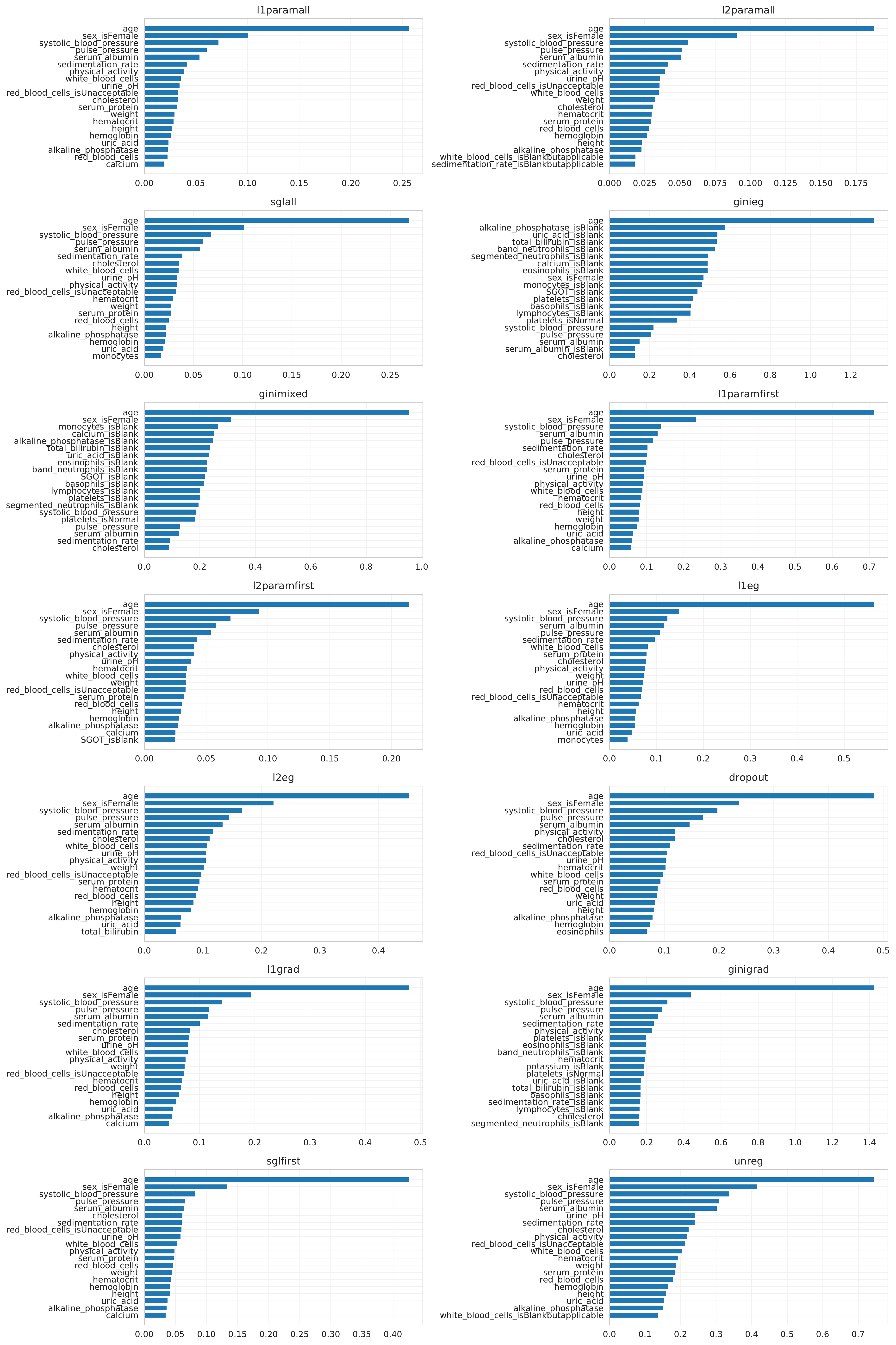}
    \caption{Summary of feature attributions for top 20 features from each model (best model from each class chosen as described in the main text); importances are an average over the 200 small-data subsamples.}
    \label{fig:shapsummaryplots}
\end{figure}

\textbf{Feature Importance Summary: }We also show summaries of the mean absolute feature importance for the top 20 features in each model in Figure \ref{fig:shapsummaryplots}.

\begin{figure}
    \centering
    \includegraphics[width=\linewidth]{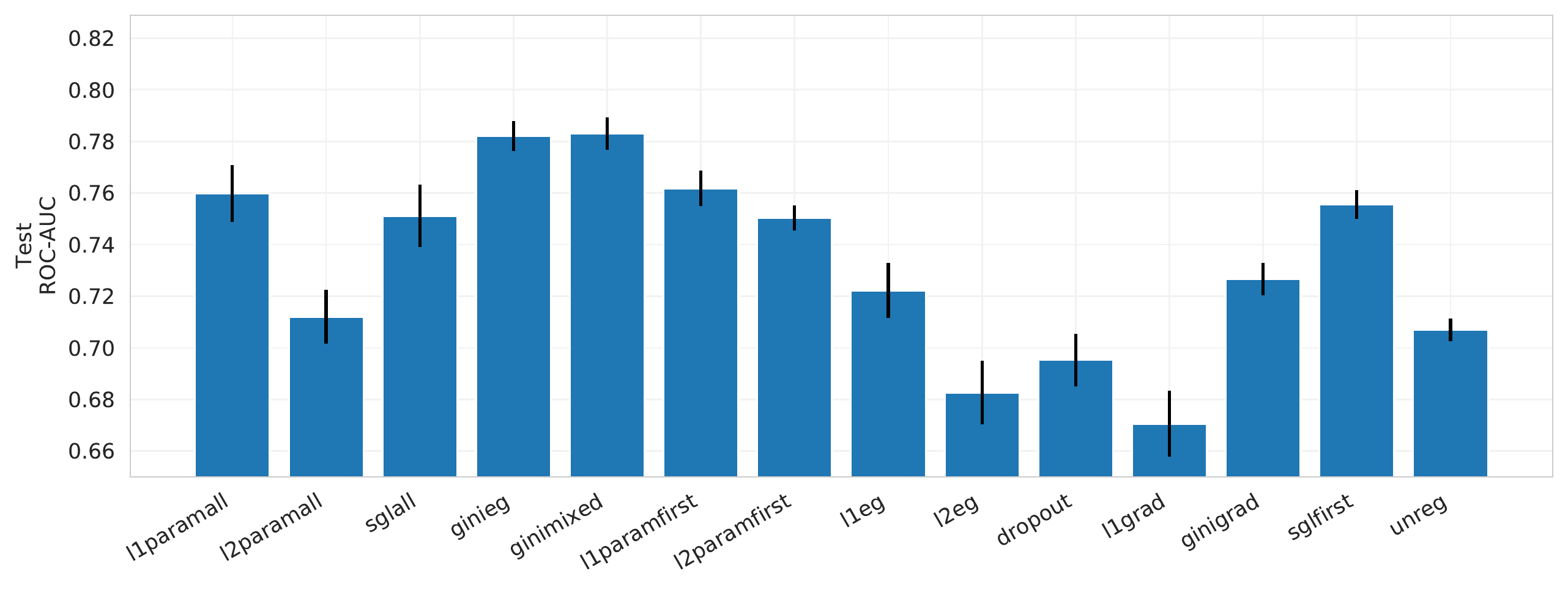}
    \includegraphics[width=\linewidth]{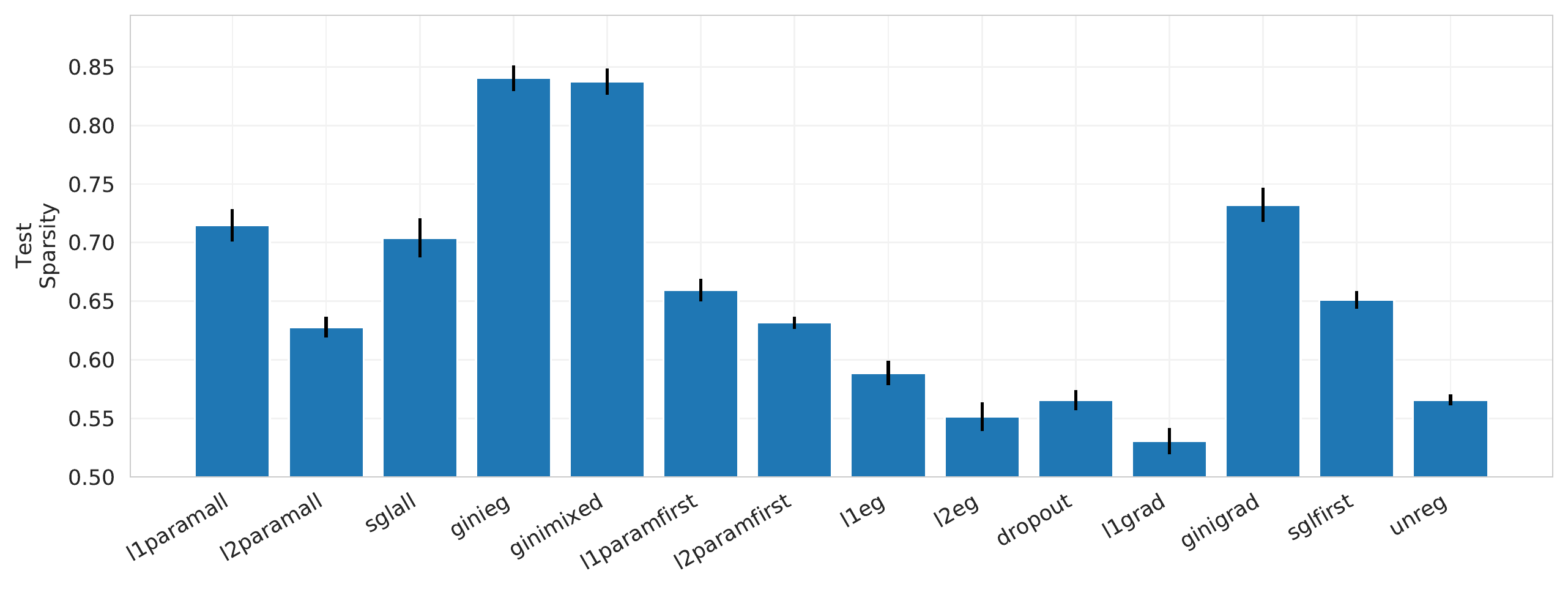}
    \caption{Performance and sparsity bar plots for all models.}
    \label{sparsevsperf}
\end{figure}

\textbf{Performance and Sparsity: }
We also display test ROC-AUC and sparsity results (with means and error bars calculated by the same method as in the maintext) for all possible penalties in Figure \ref{sparsevsperf}.

\FloatBarrier
\newpage
\printbibliography
\makeatletter\@input{gensuppaux.tex}\makeatother